%% file: main.tex
\documentclass[10pt,twocolumn,letterpaper]{article}

\usepackage[pagenumbers]{cvpr} 

\input{preamble}

\definecolor{cvprblue}{rgb}{0.21,0.49,0.74}
\usepackage[pagebackref,breaklinks,colorlinks,citecolor=cvprblue]{hyperref}
\usepackage[accsupp]{axessibility} 
\usepackage{array}
\usepackage{fancyhdr}
\usepackage{xcolor}

\newcommand{\B}[1]{\textbf{#1}}

\def\paperID{1498} 
\def\confName{CVPR}
\def\confYear{2024}

\title{AM-RADIO: \underline{A}gglomerative Vision Foundation \underline{M}odel \\ \underline{R}educe \underline{A}ll \underline{D}omains \underline{I}nto \underline{O}ne }

\author{Mike Ranzinger$^*$, Greg Heinrich$^*$, Jan Kautz, Pavlo Molchanov\\
NVIDIA\\
{\tt\small \{mranzinger,gheinrich,jkautz,pmolchanov\}@nvidia.com}
}

\begin{document}



\twocolumn[{%
\renewcommand\twocolumn[1][]{#1}%
\maketitle
\begin{center}
    \centering
    \captionsetup{type=figure}
    \includegraphics[width=0.32\textwidth,trim={0 0 0cm 0},clip]{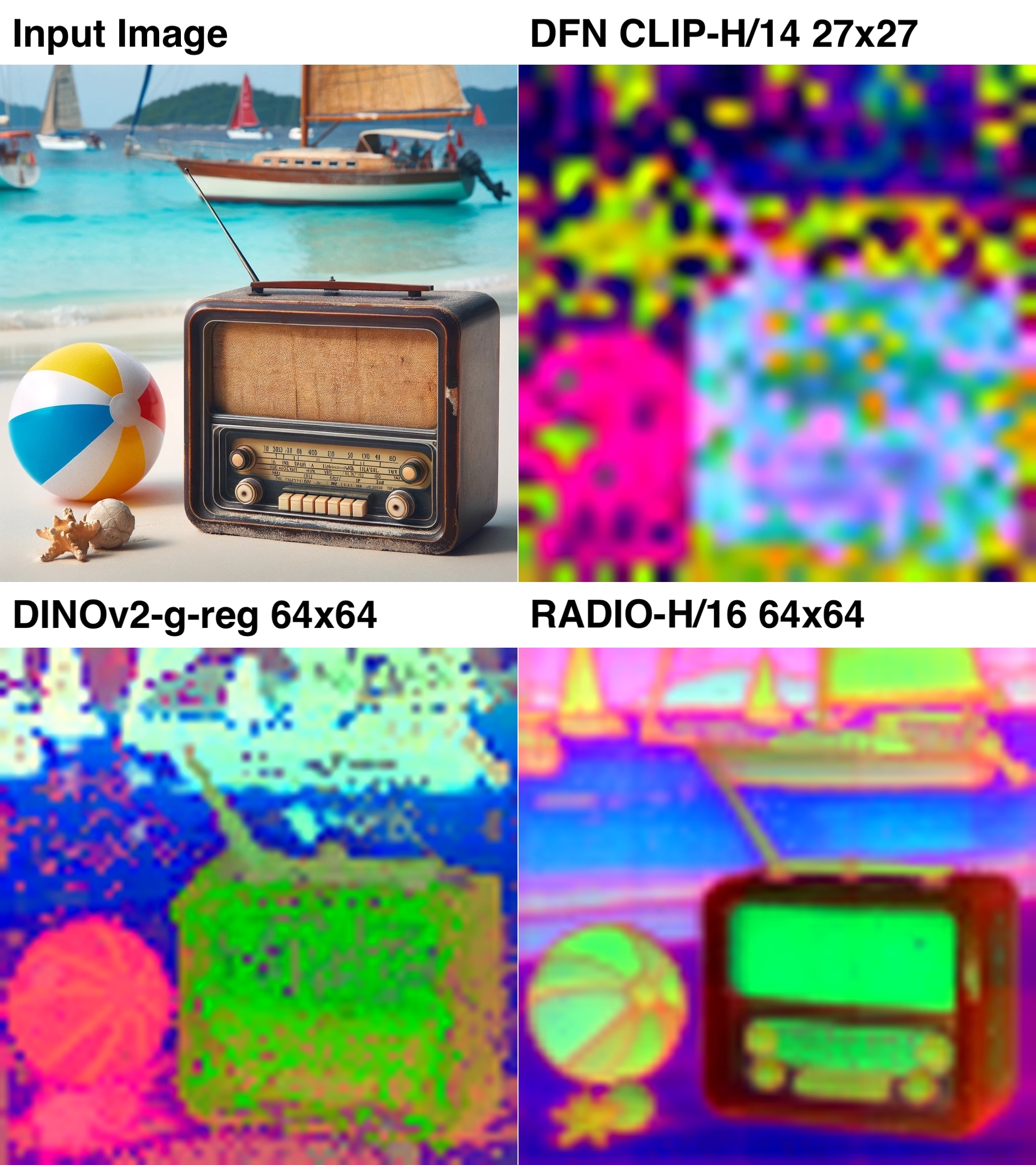}
    \quad \quad
    \includegraphics[width=0.24\textwidth]{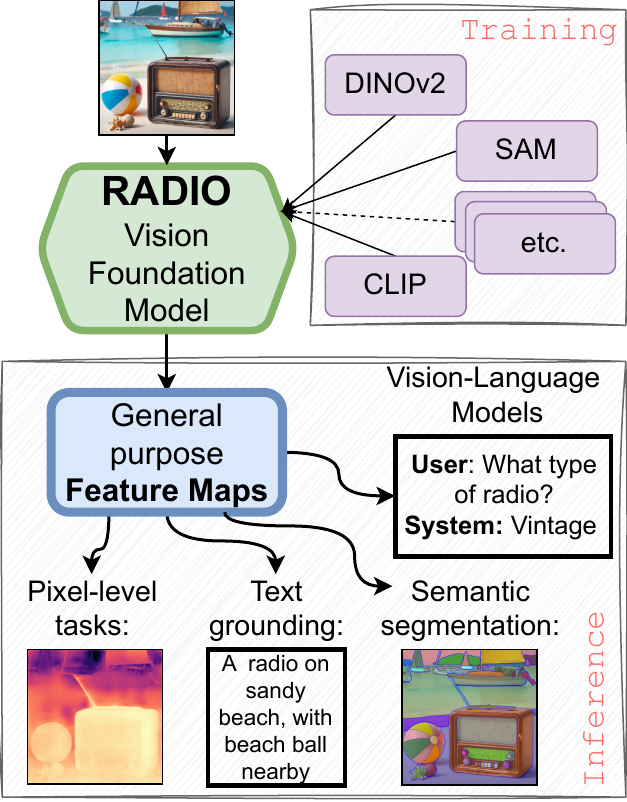}
    \quad \quad
    \includegraphics[width=0.34\textwidth]{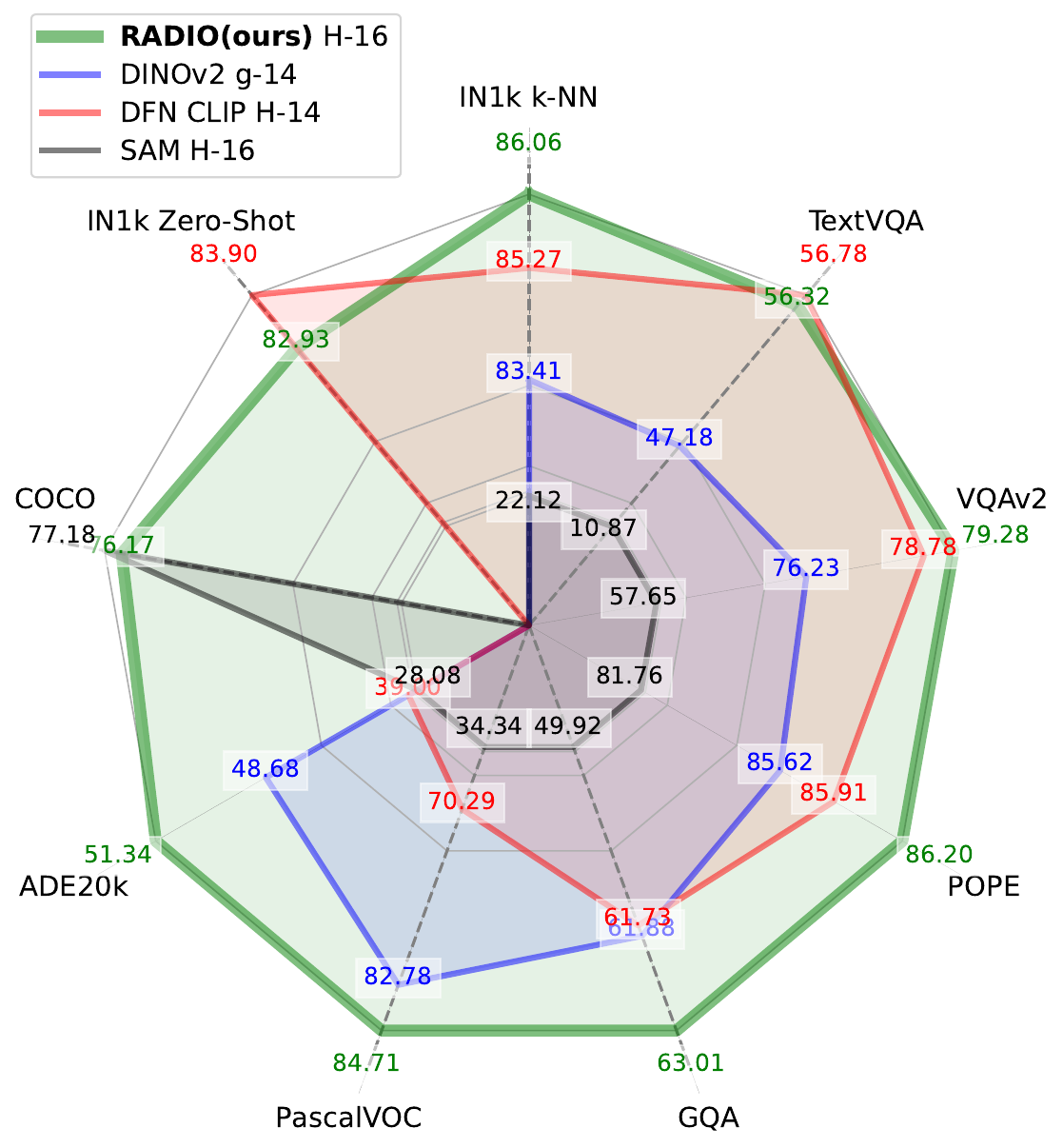}
    \captionof{figure}{
    AM-RADIO is a framework to distill multiple pretrained vision foundation models, such as CLIP~\cite{radford2021clip}, DINOv2\cite{oquab2023dinov2}, SAM~\cite{kirillov2023sam}, into a single model that we call RADIO. As a result, a single vision foundation model agglomerates unique properties of the original models. This unifying approach obtains state-of-the-art feature representations in a single forward pass while also enabling unique properties such as zero-shot classification (CLIP) or open set instance segmentation (SAM) at negligible additional cost. 
    \\
    Image description: (left) PCA feature visualization of different models. Our proposed RADIO model can process any resolution and aspect ratio, and produces semantically rich dense encodings; (middle) the overview of the AM-RADIO framework; (right) benchmarks on classification, segmentation, and vision-language modeling tasks, see section~\ref{sec:results}.  }
    \label{fig:pca_features}
    
\end{center}%
}]

\thispagestyle{fancy}
\fancyhf{}  
\fancyhead[L]{\textcolor{gray}{CVPR 2024 Conference Paper}}  
\renewcommand{\headrulewidth}{0pt}
\fancyfoot[C]{\thepage}  

\def\thefootnote{*}\footnotetext{Equal contribution}

\renewcommand{\thefootnote}{\arabic{footnote}}
\pagestyle{plain}

\input{sec/0_abstract}
\vspace{-6mm}
\input{sec/1_intro}
\input{sec/2_relatedwork}
\input{sec/3_knowledge_agglomeration}
\input{sec/5_implementation_details}

\input{sec/6_results}

\input{sec/7_conclusion}

{
    \small
    \bibliographystyle{ieeenat_fullname}
    \bibliography{main}
}

\input{sec/X_suppl}

\end{document}

%% file: preamble.tex
\usepackage[dvipsnames]{xcolor}
\usepackage{mathabx}

\usepackage{multirow}
\usepackage{listings}

%% file: sec/0_abstract.tex
\begin{abstract}
\vspace{-3mm}
A handful of visual foundation models (VFMs) have recently emerged as the backbones for numerous downstream tasks. VFMs like CLIP, DINOv2, SAM are trained with distinct objectives, exhibiting unique characteristics for various downstream tasks. We find that despite their conceptual differences, these models can be effectively merged into a unified model through multi-teacher distillation. We name this approach AM-RADIO (Agglomerative Model -- Reduce All Domains Into One). This integrative approach not only surpasses the performance of individual teacher models but also amalgamates their distinctive features, such as zero-shot vision-language comprehension, detailed pixel-level understanding, and open vocabulary segmentation capabilities. 
Additionally, in pursuit of the most hardware-efficient backbone, we evaluated numerous architectures in our multi-teacher distillation pipeline using the same training recipe. This led to the development of a novel architecture (E-RADIO) that exceeds the performance of its predecessors and is at least 6x faster than the teacher models at matched resolution.
Our comprehensive benchmarking process covers downstream tasks including ImageNet classification, semantic segmentation linear probing, COCO object detection and integration into LLaVa-1.5. 

Code: \url{https://github.com/NVlabs/RADIO}.

\end{abstract}

%% file: sec/1_intro.tex
\begin{figure}
  \centering
  \includegraphics[width=0.7\linewidth]{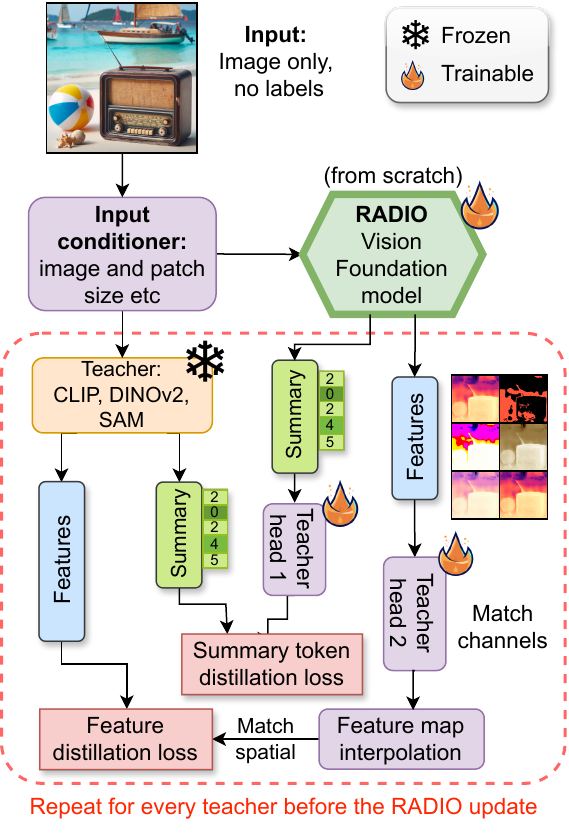}

  \caption{AM-RADIO - is a multi-teacher distillation framework that efficiently trains new vision foundation models of arbitrary architecture. It unifies unique attributes (like zero-shot text grounding, dense correspondence) of each teacher into a single model that even outperforms them on a majority of the tasks.  
  }
  \label{fig:arch_overview}
  \vspace{-4mm}
\end{figure}

\section{Introduction}
\label{sec:intro}

\begin{table*}[]
\centering\resizebox{\linewidth}{!}{
    \begin{tabular}{lcccccccccccc}
        \toprule
        \multirow{2}{*}{\textbf{Model}}                 & \B{Params} & \textbf{Resol-} & \multirow{2}{*}{\textbf{Throughput}} & \multicolumn{2}{|c|}{\B{ImageNet1K}} & \multicolumn{2}{c|}{\B{Segmentation (linear)}}   & \multicolumn{4}{|c|}{\B{Vision-Language (LLaVa-1.5~\cite{liu2023improvedllava})}} & \B{SAM~\cite{kirillov2023sam}} \\ 
                                                        &  (M)        & \B{ution}      &                & \multicolumn{1}{|c}{Zero-shot} 
                                                                                                                                   & \multicolumn{1}{c|}{k-NN}  
                                                                                                                                               & ADE20k     & \multicolumn{1}{c|}{VOC}   
                                                                                                                                                                        & GQA       & POPE      & TextVQA     & \multicolumn{1}{c|}{VQAv2} 
                                                                                                                                                                                                                          & COCO \\
        \midrule
        \midrule
        OpenCLIP-H/14 \cite{cherti2022reproducible}     & 632         & 224            & 503            & 77.19                    & 81.10     & 40.04      & 68.03     & 57.94     & 83.61     & 50.48       & 72.24     & -\\ 
        MetaCLIP-H/14 \cite{xu2023metaclip}             & 632         & 224            & 486            & 80.51                    & 82.12     & 35.39      & 62.62     & 60.57     & 84.76     & 53.65       & 75.71     & -\\
        SigLIP-L/14 \cite{zhai2023sigmoid}              & 428         & 384            & 241            & 82.61                    & 85.16     & 40.53      & 70.31     & 57.70     & 84.85     & 56.65       & 71.94     & -\\
        Intern-ViT-6B \cite{chen2023internvl}           & 5,902       & 224            & 63             &83.20${}^{\dagger\dagger}$& 78.43     & 47.20      & 76.85     & 60.18     & 84.02     & 52.45       & 76.75     & -\\
                                                        & 5,537       & 448            & 14             &${}^{\dagger\dagger}$     & 68.64     & 42.78      & 74.43     & 61.19     & \B{87.23} & \B{60.36}   & 78.83     & -\\
        *DFN CLIP-H/14 \cite{fang2023data}              & 633         & 378            & 170            & \B{83.90}                & 85.27     & 39.00      & 70.29     & 61.73     & 85.91     & 56.78       & 78.78     & -\\
        *OpenAI CLIP-L/14 \cite{radford2021clip}        & 305         & 336            & 414            & 75.54                    & 79.80     & 36.51      & 67.04     & 62.20     & 86.09     & 57.92       & 78.49     & -\\
        *DINOv2-g/14-reg \cite{darcet2023vision}        & 1,137       & 224            & 294${}^\dagger$&    -                     & 83.41     & 48.68      & 82.78     & 61.88     & 85.62     & 47.18       & 76.23     & -\\
        *SAM-H/16 \cite{kirillov2023sam}                & 637         & 1024           & 12             &    -                     & 22.12     & 28.08      & 34.34     & 49.92     & 81.76     & 43.91       & 57.65     & \B{77.18} \\
        \midrule
        \B{E-RADIO-L \textit{(Ours)}}                   & 391         & 512            & 468            & 80.73                    & 83.89     & 48.22      & 81.64     & 61.70      & 85.07      & 51.47        & 76.73      & 76.31 \\
        \B{RADIO-ViT-H/16 \textit{(Ours)}}              & 653         & 432            & 158            & 82.93                    & \B{86.06} & \B{51.34}  & \B{84.71} & \B{63.01} & 86.20     & 56.32       & \B{79.28} & 76.23 \\
        \bottomrule
    \end{tabular}
    }
    
        \caption{Comparison of vision foundation and RADIO models. ``Zero-Shot'' and k-NN are computed on ImageNet-1K. ADE20K~\cite{zhou2017ade20k} and VOC (PascalVOC2012) refer to linear probe semantic segmentation mIOU. GQA, POPE (popular), TextVQA, and VQAv2 are obtained via LLaVa 1.5~\cite{liu2023improvedllava} by replacing the vision encoder. COCO is the instance segmentation metric introduced by \cite{cai2023efficientvit} to evaluate SAM~\cite{kirillov2023sam} distillation.
        RADIO attains the best metrics on most benchmarks, and is competitive with the rest, while E-RADIO enables high quality results in resource constrained settings. Note that Zero-Shot and COCO use teacher's decoder head that is not finetuned. Throughput computed using NVIDIA A100 GPU, stated resolution, and TensorRT v8601.  *\textit{Denotes teachers used to train our final RADIO.} ${}^\dagger$\textit{We failed to export DINOv2-g-reg to TensorRT, so we report DINOv2-g here, which should be fairly close. ${}^{\dagger\dagger}$\textit{We were unable to get zero shot working using their model code.}}}
    \label{tab:teacher_metrics}
    \vspace{-4mm}
\end{table*}

Knowledge Distillation \cite{hinton2015distilling} has been a very successful and popular technique for transferring the knowledge of a ``teacher'' model (or ensemble of models) into a typically smaller ``student'' model. In the original formulation, both the student and the teacher operate on the same in-domain dataset, and the student simultaneously matches the logits of the teacher, and the ground truth labels. Instead of using labeled images, an alternative approach is to train the student model to match the features of the teacher model \cite{Romero2014FitNetsHF,huang2017like,ahn2019variational,heo2019overhaul,zagoruyko2017paying,sun2021dynamic,wei2022featuredistillation}.

Instead of using a smaller student model, \cite{xie2020noisystudent} employ an iterative learning procedure with a high-capacity model where a student of equal or greater capacity than the teacher is trained with heavy augmentation applied to the student. Once trained, they expand the dataset by pseudo-labeling new data using the trained student. They then make the student become the teacher, and repeat the process. An important finding in this work is that the student is capable of surpassing the performance of the teacher.

The authors of \cite{hinton2015distilling} explore the concept of ensemble distillation, where there are multiple teachers, each of which having restricted domain knowledge. \cite{zuchniak2023multiteacher} provides an overview of multi-teacher distillation, and proposes that instead of matching the summary of an ensemble of teachers, the student can match the features of each individual teacher via some learned non-shared mapping from the representation space of the student to each teacher. Of interest in their approach is that the student and teacher don't need to share the same architecture, and also that treating teachers individually yields improved performance.

Recently, the concept of Foundation Models (FMs) \cite{awais2023foundational} has emerged, with the general understanding that these models are large, general, and expensive to train. Through training on very large datasets they are broadly applicable to numerous downstream tasks. A seminal example of such models is CLIP \cite{radford2021clip}, which trains on web-scale weakly supervised (image, caption) pairs, and results in exceptional zero-shot performances on a wide array of computer vision benchmarks. While CLIP is firmly a FM, another model, DINOv2 \cite{oquab2023dinov2} has emerged with broad capabilities, often surpassing CLIP on dense tasks that require strong spatial features, such as ADE20k \cite{zhou2017ade20k} and Pascal VOC \cite{Everingham15}. Separately, SAM (Segment Anything) \cite{kirillov2023sam} is gaining popularity for its excellent open-vocabulary instance segmentation abilities, whose vision encoder we hypothesize has strong dense feature representations. 

We introduce AM-RADIO with the goal of learning from multiple foundational models simultaneously. We observe that, when given a student model of sufficient capacity, it is often able to exceed any of its teachers on important axes. In addition to performing well on representative foundational benchmarks, by virtue of the training framework, our student models are able to mimic their teacher models, and thus are able to perform downstream tasks that are otherwise performed by the teachers. Examples of this include CLIP-ZeroShot applications, since the language model trained by CLIP is compatible with our student, and also Segment-Anything tasks, as the student is able to replace the vision encoder and interface with the already-trained mask decoders.

We also study the effect of using a more hardware-efficient model architecture. Most works on efficiency are not directly comparable as they use different training recipes, even when evaluated on the same dataset such as ImageNet-1k, and may be over-tuned. To this end, we evaluate more than 10 promising architectures under the same training recipe for a direct comparison. We reveal that CNN-like architectures are faster but struggle to distill ViT VFMs. This led us to the development of a novel hybrid architecture, E-RADIO, that exceeds the performance of its predecessors and is at least 6x faster than teacher models at matched resolution.

\textbf{Our main contributions are as follows:}
\begin{itemize}
    \item We describe a general methodology for distilling multiple distinct foundation models into one, including models with incompatible input resolutions.
    \item We show that these student models are able to outperform their teachers on representative benchmarks.
    \item We demonstrate that these student models can either drop-in replace their teachers, or their features can be used directly in downstream applications such as providing visual encoding for LLaVA \cite{liu2023llava,liu2023improvedllava}.
    \item We benchmark a number of efficient architectures and propose a new architecture (E-RADIO) that allows for similar model quality at significant speedups.
\end{itemize}

%% file: sec/2_relatedwork.tex
\section{Related Work}\label{sec:related_work}

\noindent \textbf{Knowledge Distillation}
The underpinning of our work is based on the method of Knowledge Distillation \cite{hinton2015distilling,kim2018paraphrasing,ba2014deep,Mirzadeh2019ImprovedKD,beyer2022goodteacher} which aims to train a ``student'' model using soft targets produced by an already-trained ``teacher'' model, using the the teacher's output logits as ``soft'' labels. Alternatively, distillation can be performed using intermediate network activations \cite{Romero2014FitNetsHF,huang2017like,ahn2019variational,heo2019overhaul,zagoruyko2017paying,sun2021dynamic,wei2022featuredistillation}. In general, due to the heterogeneous nature of the different teacher foundation models that we employ, we ignore any potential labels coming from the data, and we ignore the logits of teachers, and simply opt to match the feature representations of the teachers before any task-specific processing stages.

\noindent \textbf{Multi-Teacher Distillation}
There is also a body of work that studies distilling a student model jointly from multiple teacher models simultaneously \cite{hinton2015distilling,liu2020adamultiteachr,zuchniak2023multiteacher,yuan2020reinforced,zhao2022collabteaching,yang2020modelcompression,Park2020FeatureLevelEK,you2017multiteacher,lan2018knowledge,Asif2019EnsembleKD,Fukuda2017EfficientKD}. Because of the heterogeneous domains that our teacher models cover, we don't apply approaches that marginalize teachers into a unified label, and instead map students to each teacher independently using teacher-specific projection heads from the unified student representation. Although the reason behind this method in \cite{zuchniak2023multiteacher} is different, we find the same overall strategy to be effective. While \cite{wei2022featuredistillation} doesn't study matching the features of multiple teachers simultaneously, we are able to extend their paradigm via the different projection heads. To preserve drop-in compatibility with teacher frameworks, we eliminate the feature normalization in the loss function.

\noindent \textbf{Distilling Foundation Models}
Foundation Models \cite{awais2023foundational} are meant to be generalist models that are trained on massive amounts of data, and are typically resource intensive to train from scratch. In the vein of single-teacher distillation, \cite{oquab2023dinov2} employ self-distillation to train their smaller variants from the larger teacher. \cite{wei2022featuredistillation} distills their model from a CLIP \cite{radford2021clip} teacher. Instead of focusing our energy on one teacher \textit{in particular}, we instead grab high-quality versions of CLIP \cite{radford2021clip} (using OpenCLIP \cite{ilharco2021openclip}), DINOv2 \cite{oquab2023dinov2}, and SAM \cite{kirillov2023sam}. Concurrently with our work, \cite{wang2023samclip} describe a methodology for merging a CLIP model into a pretrained SAM model via distillation, which is, in spirit, quite similar to our approach. In contrast to theirs, we include DINOv2 and also simplify the objective to straightforward feature matching. Since we don't rely on the student model to be pre-trained, it also gives us the flexibility to have the student be an architecture distinct from any teacher.

%% file: sec/3_knowledge_agglomeration.tex
\section{Knowledge Agglomeration}
\label{sec:knowledge_agglomeration}

We propose a framework to train a vision foundation model from scratch via multi-teacher distillation as shown in Figure~\ref{fig:arch_overview}. We demonstrate that each teacher brings unique properties to the foundational vision model, and the resulting trained model will agglomerate these attributes.

\subsection{Overview}
 As an initial assumption, we expect that the teacher models are capable of representing a broad swath of images found on the internet, coming from datasets such as ImageNet (1k or 21k) \cite{deng2009imagenet}, LAION-400M \cite{schuhmann2021laion400m} or DataComp-1B \cite{gadre2023datacomp}. With this in mind, we choose to study 3 seminal teacher model families: CLIP \cite{radford2021clip}, DINOv2 \cite{oquab2023dinov2}, and SAM \cite{kirillov2023sam} as they have demonstrated outstanding performance over a broad range of tasks (as in CLIP), or specifically strong performance on downstream dense tasks, such as semantic segmentation under linear probe (as in DINOv2), or open-vocabulary segmentation (as in SAM). Because these teacher models come from such diverse domains, we omit any form of supplemental ground truth guidance and treat the aforementioned datasets simply as sources of images. To assess the quality of our models, we adopt a set of representative metrics across a few broad domains.
\begin{itemize}
    \item \textbf{Image level reasoning:} (i) k-NN Top-1 accuracy on ImageNet-1K, and (ii) Zero-Shot accuracy using the CLIP teacher's language model \cite{radford2021clip}. k-NN \cite{wu2018knn,caron2021dino,oquab2023dinov2} embeds the model's summary feature vector for every image in the training set, and then for each validation image, it uses a weighted sum of the $k$ nearest training vectors to elect a label.
    \item \textbf{Pixel-level visual tasks:} segmentation mIOU on (i) ADE20K and (ii) Pascal VOC - under the linear probe setting, details in~Section~\ref{semseg}. 
    \item \textbf{Large Vision-Language Models: } we plug our frozen vision encoder model into LLaVA-1.5 \cite{liu2023improvedllava} and evaluate it on a wide set of tasks including GQA \cite{DBLP:journals/corr/abs-1902-09506}, TextVQA \cite{Singh_2019_CVPR}, ScienceQA \cite{lu2022learn} and VQAv2 \cite{balanced_vqa_v2}. Details in Section~\ref{llava}.
    \item \textbf{SAM-COCO instance segmentation:} From \cite{cai2023efficientvit}, we adopt their COCO instance segmentation methodology to evaluate our ability to replicate SAM visual features.
\end{itemize}

\noindent Results on these tasks, both for teacher models and our AM-RADIO variants, are summarized in Table \ref{tab:teacher_metrics}.

\subsection{Adaptor Heads}
We opt for simplicity in design of the adaptor heads, and leave alternative architectures as future work. To this end, we employ a simple 2-layer MLP, with a LayerNorm and GELU in between. The input dimension is the student embedding dimension, the intermediate dimension is the maximum embedding dimension of all teachers, and the output dimension matches the specific teacher. For each teacher, we employ two heads, one for the summary vector, and one for the spatial features. 

\subsection{Distillation Dataset Choice}
In table \ref{tab:dataset_choice} we study the effect of different datasets on downstream metrics. While the highest image classification metrics are achieved using ImageNet-1K as the training dataset, we argue that it doesn't fairly measure ``zero shot'' performance as the student directly learns the teacher features in the evaluation domain. For this reason, we opt for the DataComp-1B dataset.

\begin{table}[]
    \centering\resizebox{0.75\linewidth}{!}{
    \begin{tabular}{ccccc}
    \toprule
        \textbf{Dataset}      & \textbf{k-NN}  & \textbf{Zero Shot} & \textbf{ADE20K} \\
        \midrule
        ImageNet 1K           & \B{84.79}      & \B{80.44}          & 48.11  \\
        ImageNet 21K          & 84.61          & 80.10              & 48.65 \\
        LAION-400M            & 83.77          & 77.46              & 48.6 \\
        DataComp-1B           & 83.91          & 78.51              & \B{49.01} \\
        \bottomrule
    \end{tabular}
    }
    \caption{Ablation study on the choice of training dataset. We use MetaCLIP ViT-H/14~\cite{dosovitskiy2021image} and DINOv2 ViT-g/14 teachers, and a ViT-L/14 student model with CPE~\cite{kim2023regionaware}. Both ``k-NN'' and ``Zero Shot'' are for ImageNet-1k. ADE20k refers to mIOU linear probe on ADE20k.}
    \label{tab:dataset_choice}
\end{table}

\subsection{Loss Formulation} \label{lossformulation}
Because we don't have ground truth data for each teacher for each image, we instead opt to match the features coming from each teacher's vision encoder. In particular, we distinguish between the summary feature vector and the spatial feature vectors for each teacher.
The summary feature is computed differently based on the model. For CLIP and DINOv2, we use the ``class token'' as the summary feature vector, and we don't match a summary for SAM.

Let $f\left(x|\Theta_0^{}\right)$ be the student vision encoder with parameters $\Theta_0$, and $y_i^{s}=h_i^{(s)}(x'|\Theta_i^{(s)})$ be the learned student head matching teacher summary features $z_i^{(s)}=t_i^{(s)}\left(x|\Phi_i\right)$ with student adaptor parameters $\Theta_i^{(s)}$ and teacher parameters $\Phi_i$.

\begin{equation}
\begin{aligned}
    x' &= f\left(x|\Theta_0\right); & 
    y_i^{(s)} &= h_i^{(s)}\left(x'| \Theta_i^{(s)}\right); \\
    z_i^{(s)} &= t_i^{(s)}\left(x |\Phi_i\right),
\end{aligned}
\label{eq:model_outputs}
\end{equation}

\begin{equation}
\begin{aligned}
    L_{\text{summary}}(x) &= \sum_i \lambda_i L_{\text{cos}}(y_i^{(s)}, z_i^{(s)})
\end{aligned}
\label{eq:summary_loss}
\end{equation}

We found empirically that cosine distance loss produced better models compared to L1, MSE, Smooth-L1 \cite{girshick2015fast}. Additionally, supervising the spatial features of the model by matching the teacher was not only important for downstream dense tasks, but also improved the holistic quality of our model.

\begin{table}[]
    \centering
    \resizebox{0.7\linewidth}{!}{
    \begin{tabular}{l|ccc}
    \toprule
         \textbf{Teachers} & \textbf{Zero Shot} & \textbf{k-NN}  & \textbf{ADE20K} \\
         \midrule
         None              & \textbf{75.77}     & 82.59          & 41.18 \\
         CLIP              & 75.64              & 82.60          & 44.42 \\
         DINOv2            & 74.68              & \textbf{83.02} & 47.05 \\
         Both              & 74.85              & 82.96          & \textbf{48.13} \\
         \bottomrule
    \end{tabular}
    }
    \caption{Ablation over which teachers we supervise the spatial features. We use a ViT-L/14 student model and train on the LAION-400M dataset. Adding this loss term is always beneficial. DINOv2 appears to provide better spatial features than CLIP, but training the student to match both teachers produces the best results. We don't ablate SAM as we solely want it for its spatial features.}
    \label{tab:feature_distillation}
\end{table}

For matching the spatial features, we employ a combination of cosine similarity and smooth L1. Similar to equation~\eqref{eq:summary_loss} where we found that cosine similarity produced the best results, we found the same to be true for the spatial features. However, we want to allow our student model to be a drop-in replacement in the teacher frameworks, thus it's important that we match the magnitude of the teacher vectors, and so we include smooth L1. In \eqref{eq:feature_distillation} we show the formulation of this loss. Let $h_i^{(v)}(x'|\Theta_i^{(v)})$ be the learned student head for matching teacher feature vectors, and corresponding $t_i^{(v)}(x|\Phi_i^{(v)})$ be the teacher feature vectors, with $x' = f(x|\Theta_0)$, then the spatial feature loss is:

\begin{equation}
\begin{aligned}
    L_{\text{match}}(x, y) &= \alpha L_{\text{cos}}(x, y) + \beta L_{\text{smooth-l1}}(x, y) \\
    L_{\text{features}}(x) &= \sum_i \gamma_i L_{\text{match}}\left(h_i^{(v)}(x'|\Theta_i^{(v)}), t_i^{(v)}(x|\Phi_i^{(v}))\right)
\end{aligned}
\label{eq:feature_distillation}
\end{equation}

We choose $\alpha = 0.9$ and $\beta = 0.1$ to mostly rely on the empirically better cosine distance, but to also match vector magnitudes.

\subsubsection{Loss Balancing}
Due to the number of possible combinations of loss weights between the different teachers, and even which teachers, and possible formulations of loss functions, we mostly opted toward naive loss balancing with all teachers equally weighted for spatial features ($\gamma_i = 1$). For summary features, we have $\lambda_{CLIP} = \lambda_{DINO} = 1$ and $\lambda_{SAM} = 0$.

We did experiment with automatic loss balancing using predicted uncertainty \cite{cipolla2018autobalance}, AdaLoss \cite{hu2019adaloss} (momentum 0.99) and separately with AMTML-KD \cite{liu2020adamultiteachr}, as ways to learn the balance of $\lambda_i$ and $\gamma_i$. In the case of AMTML-KD, the model would always collapse its entire weight around the CLIP teacher and would yield worse results than naive manual balancing. Based on the results in table \ref{tab:loss_auto_balance}, there is very little advantage to the more exotic balancing schemes, so we opt for the "Naive" method throughout the rest of the paper. 

\begin{table}[]
    \centering

    \centering\resizebox{0.8\linewidth}{!}{
    \begin{tabular}{l|ccc}
    \toprule
         \textbf{Method}      & \textbf{Zero Shot} & \textbf{k-NN}  & \textbf{ADE20K} \\
         \midrule
         Naive                & 70.63              & 79.50          & 44.71 \\
         Uncertainty~\cite{cipolla2018autobalance}          & 70.92              & 79.37          & 44.57 \\
         AdaLoss~\cite{hu2019adaloss}              & 71.31              & 79.77          & 44.36 \\
         \bottomrule
    \end{tabular}
    }
    \caption{Loss term balancing methods comparison. We use a ViT-B/14 student, and CLIP+DINOv2 teachers. We found that AdaLoss produces the best results on the ImageNet tasks, but the worst on ADE20K.}
    \label{tab:loss_auto_balance}
    \vspace{-3mm}
\end{table}

%% file: sec/5_implementation_details.tex
\section{Implementation Details}
\label{sec:implementation}

Performing heterogeneous multi-teacher distillation is not trivial due to a mismatch in feature dimensions, input resolutions, concepts for loss computation, and downsampling ratios, as well as challenges in fitting multiple teachers into a single GPU.

\noindent \textbf{General.}
We train all student models using the AdamW \cite{loshchilov2018adamw} optimizer, batch size 1024, cosine annealing learning rate schedule and base learning rate of $0.001$. We train for 600k steps, resulting in 614M total examples seen. For our best student model, we train using DFN CLIP ViT-H/14 378px, OpenAI CLIP ViT-L/14 336px, DINOv2 ViT-g/14 224px, and SAM ViTDet-H 1024px. We apply random scale + cropping to both student and teacher inputs. We chose the DataComp-1B dataset due to it having the highest quality results of the web-scale datasets we had access to. We train in two stages, first with CLIP+DINOv2 for 300k steps at 256px, and second with CLIP+DINOv2 at 432px plus SAM at 1024px for 300k steps.

\noindent \textbf{Student architecture.}
We study two settings for student model architecture:
\begin{itemize}
\item Standard ViT~\cite{dosovitskiy2021image} architecture to match the architecture of teachers. Our best model is a ViT-H/16.

\item Efficient architecture variants prioritizing high throughput on GPUs. See~Section~\ref{sec:efficient}.

\end{itemize}

\noindent \textbf{Multi-scale Teachers.}
We choose ViT-H/16 architecture for our student model. To match resolution of SAM features, we feed the expected resolution of $1024^2$. Given that our CLIP and DINOv2 teachers are patch-14 models, we opt to feed the student $432^2$ inputs, as that is the same effective resolution as $378^2$ for patch-14. We found that interpolating DINOv2 features doesn't degrade results, so the teacher operates at 224px and we upsample the outputs to match the student.

\noindent \textbf{Rank/Teacher Partitioning.}
We group teacher models by (batch\_size, student\_resolution), and then distribute the groups to different GPUs, such that each GPU processes a consistent batch size and input resolution. We also sample groups at different rates. For our training setups that include SAM, we train with 64 GPUs, half of which get the CLIP+DINOv2 group with batch size 32 per GPU and input resolution 432, and the other half get SAM with batch size 2 per GPU and input resolution 1024. This results in an effective batch size of 1,152. For CLIP+DINOv2 training, we use 32 GPUs, resulting in batch size 1024.

\begin{figure}[h]
    \centering
    \begin{subfigure}[b]{0.2\textwidth}
        \centering
        \includegraphics[width=\textwidth]{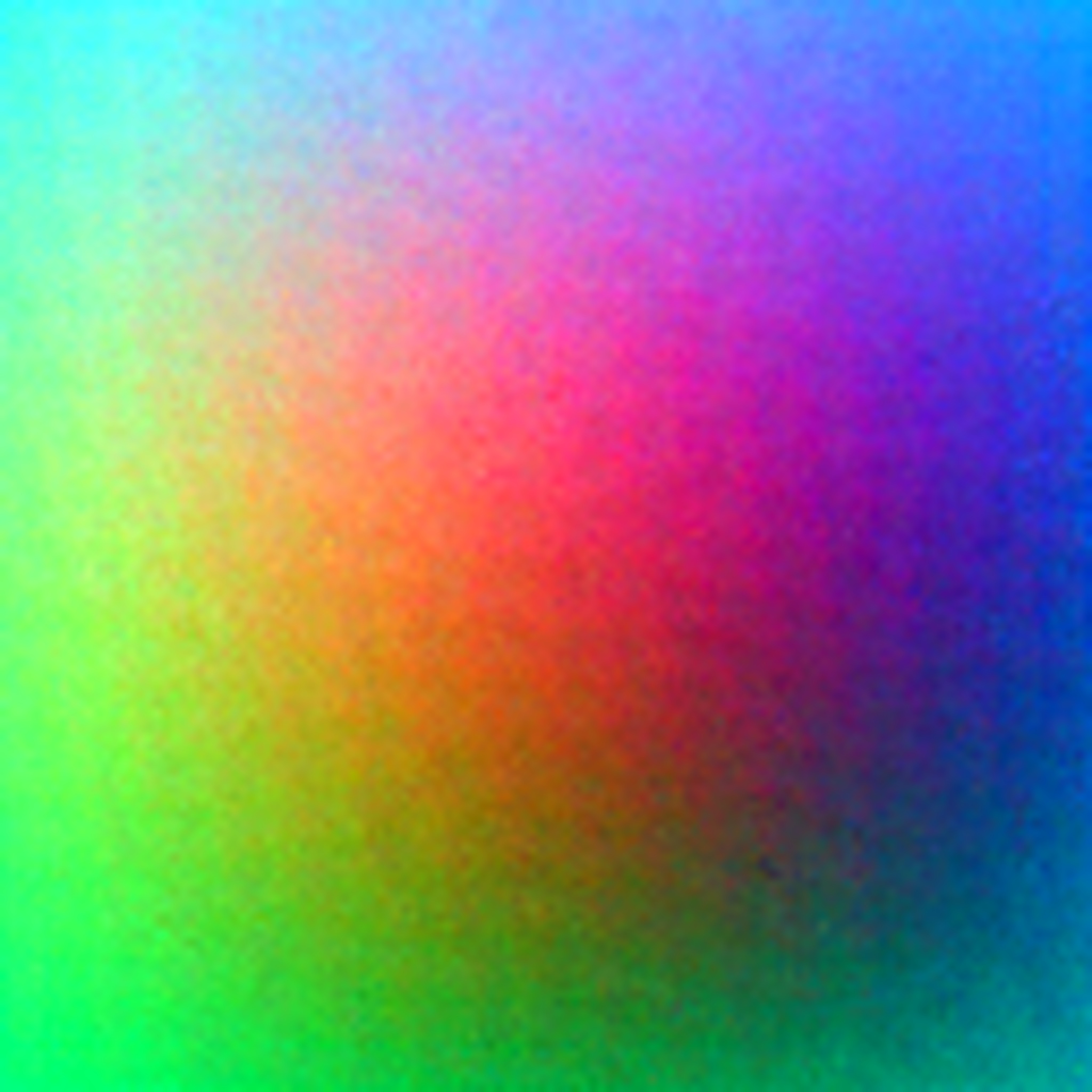}
        \caption{RADIO 2048px}
        \label{fig:sub1}
    \end{subfigure}
    \begin{subfigure}[b]{0.2\textwidth}
        \centering
        \includegraphics[width=\textwidth]{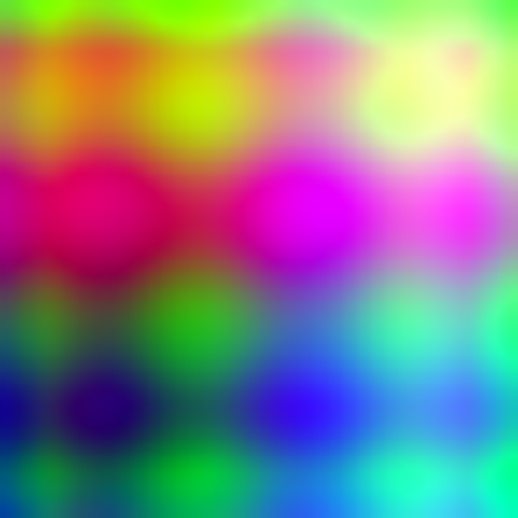}
        \caption{DINOv2-g-reg 518px}
        \label{fig:sub2}
    \end{subfigure}
    \begin{subfigure}[b]{0.2\textwidth}
        \centering
        \includegraphics[width=\textwidth]{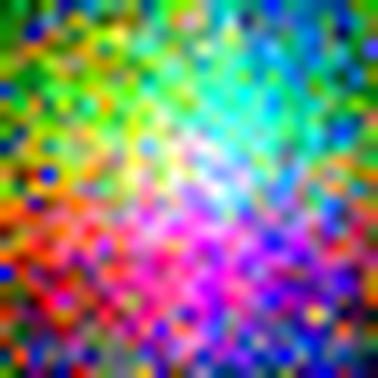}
        \caption{DFN CLIP 378px}
        \label{fig:sub3}
    \end{subfigure}
    \begin{subfigure}[b]{0.2\textwidth}
        \centering
        \includegraphics[width=\textwidth]{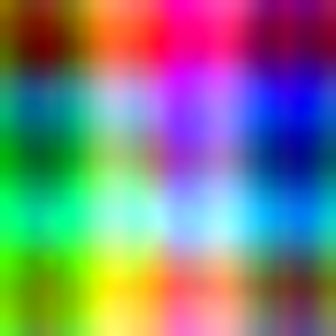}
        \caption{OpenAI CLIP 336px}
        \label{fig:sub4}
    \end{subfigure}
    \caption{PCA visualization of the position embeddings for various models. The CPE method not only allows RADIO to learn an arbitrarily large absolution position embedding map, but also goes a long way towards regularizing the space and eliminating high frequency artifacts. As seen with the other models, position embeddings normally have regular frequency patterns, leading to undesirable output artifacts from the ViT \cite{yang2024denoising,yang2023emernerf,bolya2023window}.}
    \label{fig:pos_embed_pca}
\end{figure}

\noindent \textbf{Multi-Resolution ViTs.}
Many of our student models use ViT \cite{dosovitskiy2021image} as the base vision architecture. Traditionally, ViTs use a learned position embedding for each input patch in an image, which in turn enforces that the model always operates at a constant resolution. We employ the Cropped Position Embedding (CPE) \cite{kim2023regionaware} augmentation with the number of positions being equal to $128^2$. The position embeddings are then randomly cropped and interpolated to match the number of input patches for the student model. Even when training with CLIP+DINOv2 at 224 resolution, we found that this technique results in a negligible drop (Table \ref{tab:cpe_comparison}) in summary metrics, but \textit{improved} semantic segmentation linear probing mIOU. 
For heterogeneous-resolution students, this is a seamless technique that allows ViT to operate at arbitrary resolutions within some envelope. In addition to enabling arbitrary resolutions, as shown in figure \ref{fig:pos_embed_pca}, CPE reduces the noise artifacts in the position embeddings as compared to other ViT models \cite{yang2024denoising,yang2023emernerf,bolya2023window}.

\begin{table}[]
\centering
    \resizebox{0.47\linewidth}{!}{
    \begin{tabular}{lcc}
         \toprule
         \textbf{Method}                              & \textbf{k-NN}  & \textbf{ADE20K} \\
         \midrule
         Non-CPE                                         & \B{82.96}     & 47.30 \\
         CPE                                            & 82.84          & \B{48.52} \\
         \bottomrule
    \end{tabular}
    }
    \caption{Comparing identical ViT-L/14 student models, with and without CPE \cite{kim2023regionaware} formulation. While the student only ever trains at $224^2$ resolution, CPE allows us to generalize to $518^2$ resolution, not only improving over non-CPE, but even outperforming DINOv2-g itself.}
    \label{tab:cpe_comparison}
    \vspace{-4mm}
\end{table}

\noindent \textbf{High-Resolution ViT Student.}
In SAM, they employ the ViTDet \cite{li2022vitdet} architecture as a way to reduce the computational and memory burden of ViT models at high-resolution. We reformulate this arch instead into a training augmentation, where we sample a window size from a set of possible window sizes. This allows us to reduce the computational burden of training the student model with the SAM teacher, and, as we make the window size flexible, it provides an additional throughput scaling mechanism during inference. Table \ref{tab:sam_coco_results} demonstrates our ability to replace SAM's encoder. Separately, we found that high resolution training was unstable, so we apply spectral reparametrization \cite{zhai2023stabilizing} and a weight decay of $0.02$ to prevent attention entropy collapse.

\noindent \textbf{Student/Teacher Resolution Mismatch.}
When the student and teacher downsample images through their processing stack at different rates, it results in the output feature vectors having different resolutions. For example, if the teachers use a ViT-H/14 architecture and student a ViT-H/16, it means that the student outputs a $14^2$ feature map, and the teachers a $16^2$ feature map. For $L_{\text{features}}$ we bilinearly interpolate the outputs to match the larger resolution between the student and teacher features.

\begin{table}[]
    \centering
    \resizebox{0.95\linewidth}{!}{
    \begin{tabular}{rcccccccccc}
    \toprule
         \textbf{}        & \B{Zero Shot} & \B{k-NN}  & \B{ADE20K} & \B{VOC}   & \B{VQAv2} \\
         \midrule
          CLS token & 78.55         & \B{83.91} & \B{49.01}  & \B{83.51} & 77.66     \\
          Avgpool   & \B{80.12}     & 83.83     & 38.36      & 77.04     & \B{78.28} \\          
          \bottomrule
    \end{tabular}
    }
    \caption{Comparing identical ViT models, with CLS token and average pooling summarization. 
    }
    \label{tab:summarization_comparison}
    \vspace{-5mm}
\end{table}

\noindent \textbf{Feature Summarization.}
In \ref{lossformulation} we explained how teacher summary features are extracted using the ``class token'' of their respective ViT models. We now turn our attention to the summarization of student features. 
ViTs have 2 options: (i) a separate summarization ``CLS'' token or (ii) average pooling patch tokens. We evaluate both options in Table~\ref{tab:summarization_comparison}. We observe that average pooling improves summary loss, but has a more significant detrimental effect on the feature loss. Given the importance of the latter we choose to use separate CLS tokens.

%% file: sec/6_results.tex
\section{Results}
\label{sec:results}
In this section, we analyze models obtained with the proposed AM-RADIO framework. First, we touch upon backbone efficiency, then compare with the original teachers (CLIP, DINOv2, SAM), and benchmark models under vision question answering in the LLaVa framework. We will see that the proposed models outperform the original teachers in multiple metrics, including throughput. Results are shown in Figure~\ref{fig:pca_features} and Table~\ref{tab:teacher_metrics}.

\begin{figure*}
  \centering\resizebox{1.0\linewidth}{!}{

    \includegraphics{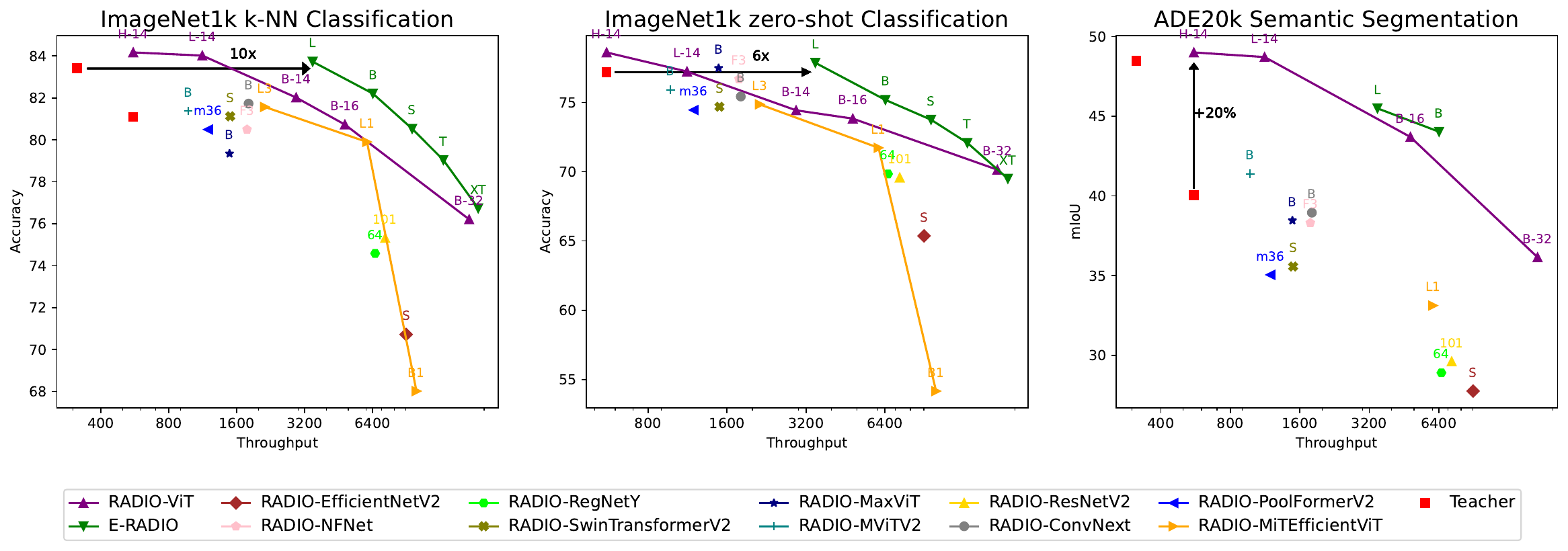}
  
  
  }
  \caption{
  All models followed the same training protocol. The results from three benchmarks show that RADIO and E-RADIO models outperform others in efficiency. This under-performance in other models might be due to overfitting architectures on supervised ImageNet-1K training. E-RADIO notably delivers results 10 times faster and with a 20\% improvement over teacher models. We study E-RADIO at 224px resolution, with a window size of 7.
 }
  \label{fig:speed_acc}
\end{figure*}

\subsection{Efficient Students}

\label{sec:efficient}

\begin{table}[]
    \centering
    \centering\resizebox{1.0\linewidth}{!}{
    \begin{tabular}{lcccccc}
    \toprule
        {Backbone}   & {Param.}        & {Through-}                  & {Zero} & {k-NN} & {ADE20k} & FD loss \\
         & Count & put & Shot & & \\
        \midrule
        \multicolumn{7}{c}{Teachers} \\
        \midrule
        DINOv2 G/14 & 1.14B & 313 &  N/A  & 83.41 & 47.53 \\
        OpenCLIP H/14 & 632M &  556 & 77.19 & 81.10 & 40.04 \\
        \midrule
        \multicolumn{7}{c}{Existing Efficient Models} \\
        \midrule
              
        \textbf{EfficientNetV2-S} & 21M & 9017 & 65.37 & 70.72  & 27.75 & 0.415 \\
        \textbf{ResNetv2-101} & 44M & 7283 & 69.58 & 75.32 & 29.61 & 0.405 \\
        RegNetY-064 & 30M & 6573 &  69.84  & 74.59 & 28.9 & 0.394 \\
        EfficientViT-L1 & 38M & 6048 & 71.73 & 79.90 & 33.12 & 0.376 \\
        ConvNext-B & 88M & 1805 & 75.43 & 81.73 & 38.95 & 0.358 \\
        NFNet-F3 & 254M & 1777 &  76.93 & 80.50 & 38.31 & 0.340 \\
        SwinV2-S & 49M & 1497 & 74.70 & 81.12 & 35.57 & 0.364 \\
        MaxViT-B & 119M & 1486 & 77.49 & 79.34 & 38.46 & 0.340 \\
        PoolformerV2-M36 & 56M & 1194 & 74.46 & 80.49 & 35.05 & 0.377 \\
        MViTV2-B & 51M & 975 & 75.92 & 81.39 & 41.39 & 0.345 \\
        \midrule
        \multicolumn{7}{c}{Proposed architecture} \\ 
        \midrule
        \textbf{E-RADIO-B} & 118M & 6422 & 75.19 & 82.21 & 44.03 & 0.319 \\ 
        \ \ $\drsh$ w/o upsample & 113M & 7040 & 75.45 & 82.05 & 41.26 & 0.353 \\
        \textbf{E-RADIO-L} & 265M & 3472 & 77.87 & 83.73 & 45.5 & 0.265 \\ 
        \bottomrule
    \end{tabular}
    }
    \caption{Comparison of backbones. Throughput is measured using TensorRT 9.0.1 on A100 in mixed FP16/FP32 precision at batch size 128 on $224^2$px resolution. Sorted by descending throughput order. FD loss is the Feature Distillation training loss against the DINOv2 teacher, it exhibits high correlation with the ADE20k mIoU. Bolded models form the speed/quality Pareto front.
    }
    \label{tab:backbones}
\end{table}

We aim to find an efficient model architecture to speed up the inference of VFM. There are a number of architectural designs aimed at high throughput on GPU devices. We use our distillation framework to evaluate several backbones with no change in training hyperparameters.

Upon reviewing the literature on efficient vision backbones focused for high GPU throughput, we pick the following list of architectures: EfficientNetV2 \cite{DBLP:journals/corr/abs-2104-00298}, ResNetv2 \cite{DBLP:journals/corr/SzegedyIV16}, RegNetY \cite{radosavovic2020designing}, FasterViT \cite{hatamizadeh2023fastervit}, EfficientViT \cite{cai2023efficientvit}, ConvNext \cite{liu2022convnet}, NFNet \cite{brock2021highperformance}, SwinV2 \cite{liu2022swin}, MaxViT \cite{tu2022maxvit}, PoolformerV2 \cite{yu2022metaformer} and MViTV2 \cite{li2022mvitv2}. We train all the backbones via distillation on the ImageNet-21k dataset, using OpenCLIP ViT-H/14 (laion2B-s32B-b79K) and DINOv2 g/14 as teachers. Results are compiled in Table \ref{tab:backbones}. 

We observe that many models lag behind teachers. Additionally, CNN-like models are significantly faster than ViTs, while the latter are more accurate. The relatively low performance of existing efficient backbones on the dense ADE20k segmentation task is not unexpected since all of them apply a spatial dimension reduction factor of 32 for final feature maps of size $7^2$ for input resolution of $224^2$px, thus hardly capable of capturing fine-grain spatial information.

\noindent \textbf{E-RADIO:} To overcome this issue, we propose a novel hybrid architecture, named E-RADIO (Efficient RADIO). This design borrows ideas from existing literature and includes an input stem with strided convolutions to downsample the input image by 4x. It then proceeds with 2 stages of YOLOv8 C2f convolution blocks and 2 stages of transformer.
For the transformer variant we pick windowed attention (like in SWIN~\cite{liu2022swin}), and interleave local windowed attention with ``global'' windowed attention as done in~\cite{hatamizadeh2023fastervit} and ViTDet \cite{li2022vitdet}. To perform ``global'' attention we first downsample the feature map by 2x, apply windowed attention, and then upsample the feature maps back to the original resolution. Up-/down-sampling is performed by strided convolution with a kernel size 3x3 and a stride of 2. The last idea is borrowed from EdgeViT~\cite{edgevit}, which uses local-global-local attention. See Appendix for details. Finally, E-RADIO upsamples final feature maps by 2x via a deconvolutional layer and adds them to feature maps from the third stage, resulting in only a 16x spatial resolution reduction. Such upsampling gives an improvement in dense task while being only 10\% slower. Results of E-RADIO in Table~\ref{tab:backbones} demonstrate that the proposed architecture significantly outperforms the competition, and can be seen as an efficient replacement for the much slower full ViT.

\subsection{Comparison with teachers}

A comprehensive set of results is presented in Table \ref{tab:teacher_metrics}. We notice that MetaCLIP is better than OpenCLIP, and DFN CLIP better than MetaCLIP. DINOv2 provides important properties for dense tasks: ADE20k and VOC. Our E-RADIO-L model is significantly faster than all ViT models. At the same time, it strongly outperforms MetaCLIP on most metrics at matched throughput, while also enabling Zero-shot capability that is absent in DINOv2 and SAM. Our full model, ViT-H/16, is as fast as the teachers but outperforms them on 6 out of 9 tasks, demonstrating the efficiency of the proposed distillation framework.

\noindent \textbf{Drop-In SAM Replacement.}
Following \cite{cai2023efficientvit}, we use their evaluation harness to compute the mIOU for instance segmentation using pretrained SAM with vision encoder replaced by our model. Table \ref{tab:sam_coco_results} shows the results of the COCO Instance Segmentation task using the baseline SAM models and RADIO. 

\begin{table}[]
    \centering
    \centering\resizebox{0.9\linewidth}{!}{
     
    \begin{tabular}{lccc}
    \multicolumn{4}{c}{COCO 2017 drop-in SAM replacement at 1024x1024} \\
    \toprule
        \textbf{Family}                   & \textbf{Arch}             & \textbf{mIOU} & \textbf{Throughput} \\
        \midrule
        \multirow{3}{*}{SAM}              & Base                      & 75.78         & 50.94 \\
                                          & Large                     & 77.02         & 20.62 \\
                                          & Huge                      & 77.18         & 11.83 \\
        \hline
        E-RADIO (ours)                    & Large                     & 76.31         & 121.74 \\
        \hline
        \multirow{2}{*}{RADIO (ours)}     & ViTDet-H/16-W8$^\dagger$  & 76.09         & 29.09 \\
                                          & ViTDet-H/16-W16$^\dagger$ & 76.23         & 27.91 \\

    \bottomrule
    \end{tabular}
    }
    
    \caption{We substitute SAM's vision encoder with our RADIO model. RADIO aligns with SAM's features just before the encoder's neck layer. We also examine the impact of varying ViTDet window sizes. Differences in throughput owe to the fact that RADIO doesn't use relative positional embeddings and we reduced shuffling with our patch reordering algorithm (in appendix). Throughput is computed on an NVIDIA A100 GPU using TensorRT and batch size 16. ${}^\dagger$This is the same model, just with a different window size setting. 
    }
    \label{tab:sam_coco_results}
\end{table}

\subsection{Semantic Segmentation Linear Probing} \label{semseg}

We train a linear head on top of the frozen features of the teachers and students alike and evaluate performance in the MMSeg \cite{mmseg2020} framework using the mIoU metric on ADE20k and PascalVOC2012 datasets. We use a training and evaluation crop size of 512 for RADIO, 518 for DINOv2, and the native resolution for the others. We use the ``slide'' evaluation mode with a stride of $\frac{2}{3}$ the crop size. We train the linear head for 160k steps using a total batch size of 16, a base learning rate of $10^{-3}$ and the AdamW optimizer.

\subsection{Visual Question Answering} \label{llava}

We replace the vision encoder in a LLaVA 1.5\cite{liu2023improvedllava} setup with our own encoder. A 2-layer MLP is used to project frozen visual features into the language token space. Under the default LLaVA 1.5 settings, we pretrain a multimodal projection MLP and then run instruction tuning to finetune a Vicuna 7B-1.5 model\cite{zheng2023judging}. We evaluate models using the validation sets of GQA \cite{DBLP:journals/corr/abs-1902-09506}, TextVQA \cite{Singh_2019_CVPR}, POPE \cite{Li-hallucination-2023} (popular), and we score the model on the Test-Dev set of VQAv2 \cite{balanced_vqa_v2} using EvalAI\cite{yadav2019evalai}. We use the vision encoder's native input resolution, resizing the long edge and padding the short edge. Experimental results are compiled in Table \ref{tab:teacher_metrics}. Owing to the increased input resolution flexibility of RADIO, we resize the long edge of the image to 432px aspect preserving, only padding to the nearest multiple of the patch size. This results in $~462$ tokens on average, versus the $576$ tokens required by the 336px patch-14 encoders, a 20\% reduction.
 
\subsection{3D Awareness Probing} \label{probe3d}

Following the work from \cite{elbanani2024probing}, we probe our model's ability to extract 3D features such as depth, surface normals and multi-view keypoint correspondance. Our results are summarized in Table \ref{tab:probe3d} and show that our model's performance is on par with the bigger DINOv2-g-14-reg\cite{darcet2023vision} and significantly better than other comparably-sized teachers.

\begin{table}[]
    \centering
    \centering\resizebox{0.9\linewidth}{!}{
     
    \begin{tabular}{lccc}
    
    \toprule
        \textbf{Backbone}       & \textbf{Depth}  &     \textbf{Surface}         & \textbf{Multi-view} \\
                                &  & \textbf{Normals} & \textbf{corr.} \\
        \midrule        
        DFN CLIP-H/14         & 52.5  & 23.0   & 20.3 \\
        OpenAI CLIP-L/14      & 53.7  & 25.3   & 20.7 \\
        DINOv2-g/14-reg       & 83.2  & 59.6   & 59.9 \\
        SAM-H/16              & 68.2  & 50.3   & 45.3 \\
        RADIO-ViT-H/16 (ours) & 81.0  & 58.5   & 62.1 \\

    \bottomrule
    \end{tabular}
    }
    
    \caption{Probing 3D Awareness: we use the code from \cite{elbanani2024probing} and evaluate our RADIO model and its teachers on monocular depth, surface normals and multi-view correspondance tasks, using the NAVI\cite{jampani2023navi} dataset. For each task we report the accuracy, averaged over all thresholds.
    }
    \label{tab:probe3d}
\end{table}

%% file: sec/7_conclusion.tex
\section{Conclusion and Key Insights}\label{sec:conclusion}

\begin{figure}
  \centering
  \begin{tabular}{c}
      \includegraphics[width=\linewidth]{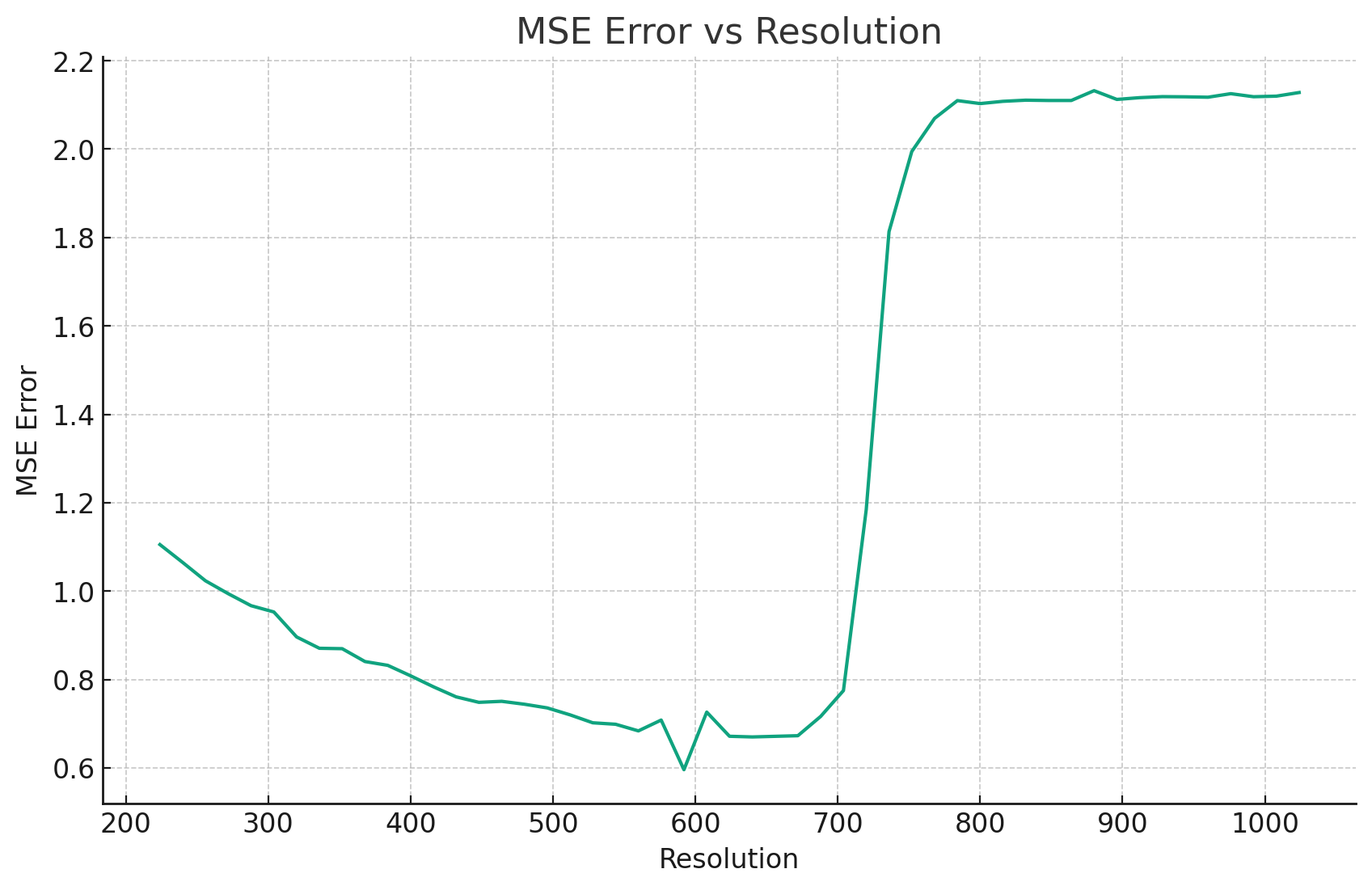} \\
      \includegraphics[width=\linewidth]{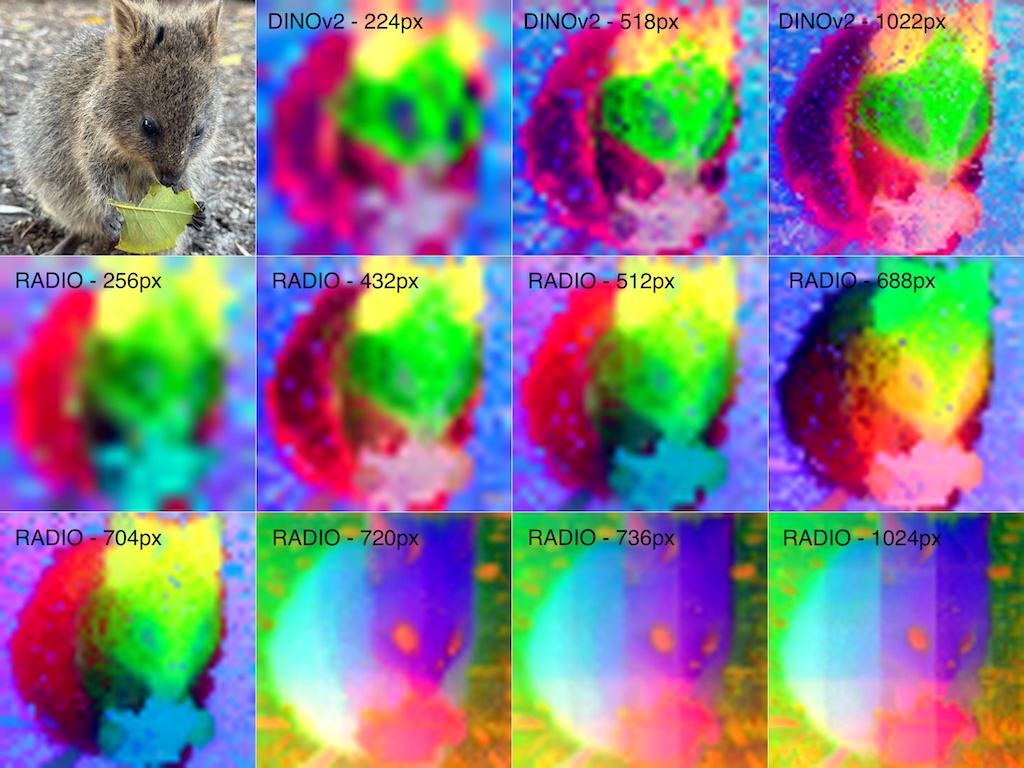}
  \end{tabular}
  
  \caption{RADIO ``mode switches'' when resolution is increased. In the plot, we show the MSE error between the RADIO features coming from its DINOv2 head at different resolutions, versus the features actually produced by DINOv2 at 518px. We bilinearly interpolate the RADIO features to match the DINOv2 feature resolution. At 720px, there is a sudden jump in the error, which corresponds with a complete change in color space in the image.}
  \label{fig:mode_switch}
  \vspace{-0.5cm}
\end{figure}

Most VFMs have unique properties such as language grounding (CLIP), dense correspondences (DINOv2), and detailed segmentation (SAM), but also large holes in capability. Distillation allows uniting all these properties in a single model that often outperforms any of the teachers. We have also observed that better teachers yield better students, which allows RADIO to absorb and challenge the current SOTA foundation models at a given point in time.

\noindent \textbf{Feature distillation loss.}
We observe the crucial importance of full feature distillation to boost the performance of the teacher in dense image understanding tasks, such as an 18\% relative improvement on ADE20K.

\noindent \textbf{SAM vs DINOv2.} We find that, out of the box, SAM is not well-suited for downstream tasks, whereas DINOv2 significantly outperforms in zero- and few-shot tasks. For example, ADE20K segmentation via linear probing is 1.7x better with the latter, and the ImageNet1k k-NN metric is 4x better. SAM excels in detecting edges and segmenting objects but performs poorly in high-level object description and combining the semantics of multiple objects (Figure~\ref{fig:speed_acc}).

\noindent \textbf{Dense features.} As seen in figure \ref{fig:pca_features}, RADIO is capable of producing high resolution and low-noise features. An issue we identified, however, shown in figure \ref{fig:mode_switch} is that RADIO appears to have a latent `low resolution' and `high resolution' mode, likely due to the partitioned training between CLIP+DINO and SAM objectives, which we intend to fix in future work.

\noindent \textbf{Efficient backbone.} Based on our analysis of distilling efficient backbones, we conclude that most model designs are overly tailored towards supervised training on ImageNet1K, and as a result, do not scale well to VFM settings. We designed a new vision backbone, E-RADIO, with a hybrid CNN-Transformer architecture that improves upon the Pareto frontier.

%% file: sec/X_suppl.tex
\clearpage
\setcounter{page}{1}
\maketitlesupplementary

\onecolumn

\appendix

\section{E-RADIO architecture details}
\begin{figure}
    \centering
    \includegraphics[width=0.9\linewidth]{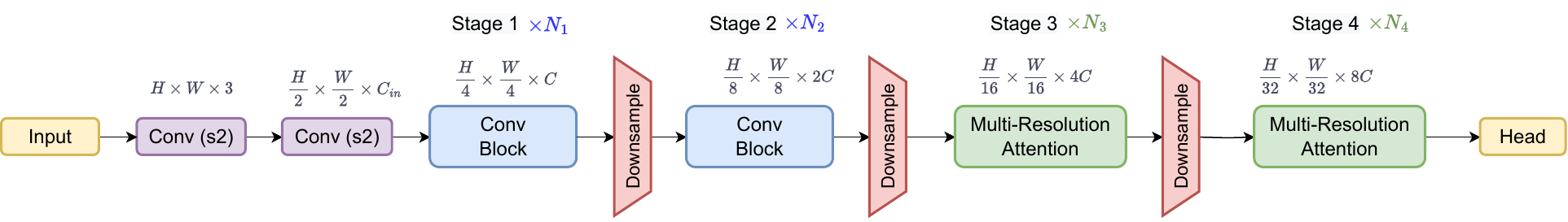}
    \caption{High level architecture of the ERADIO network architecture. Overall architecture is composed of multiple stages: 1) the stem, 2) 2 convolutional blocks from YOLOv8, 3) 2 transformer blocks with multi-resolution windowed self attention.}
    \label{fig:eradio_arch}
\end{figure}

The architecture of E-RADIO is illustrated in Figure~\ref{fig:eradio_arch}. It is a hybrid CNN-Transformer architecture. First 2 stages follow convolution paradigm and have the C2f architecture from YOLOv8 model~\cite{yolov8_ultralytics}. The last 2 stages have the Transformer architecture with windowed attention and multi-resolution attention (MRA) structure. Every stage, except the last one, are followed by downsample block. We implement it as a strided convolution with 3x3 kernel and stride 2, followed by batch normalization layer. 

\subsection{Multi-Resolution Attention}

\begin{figure}
    \centering
    \includegraphics[width=0.7\linewidth]{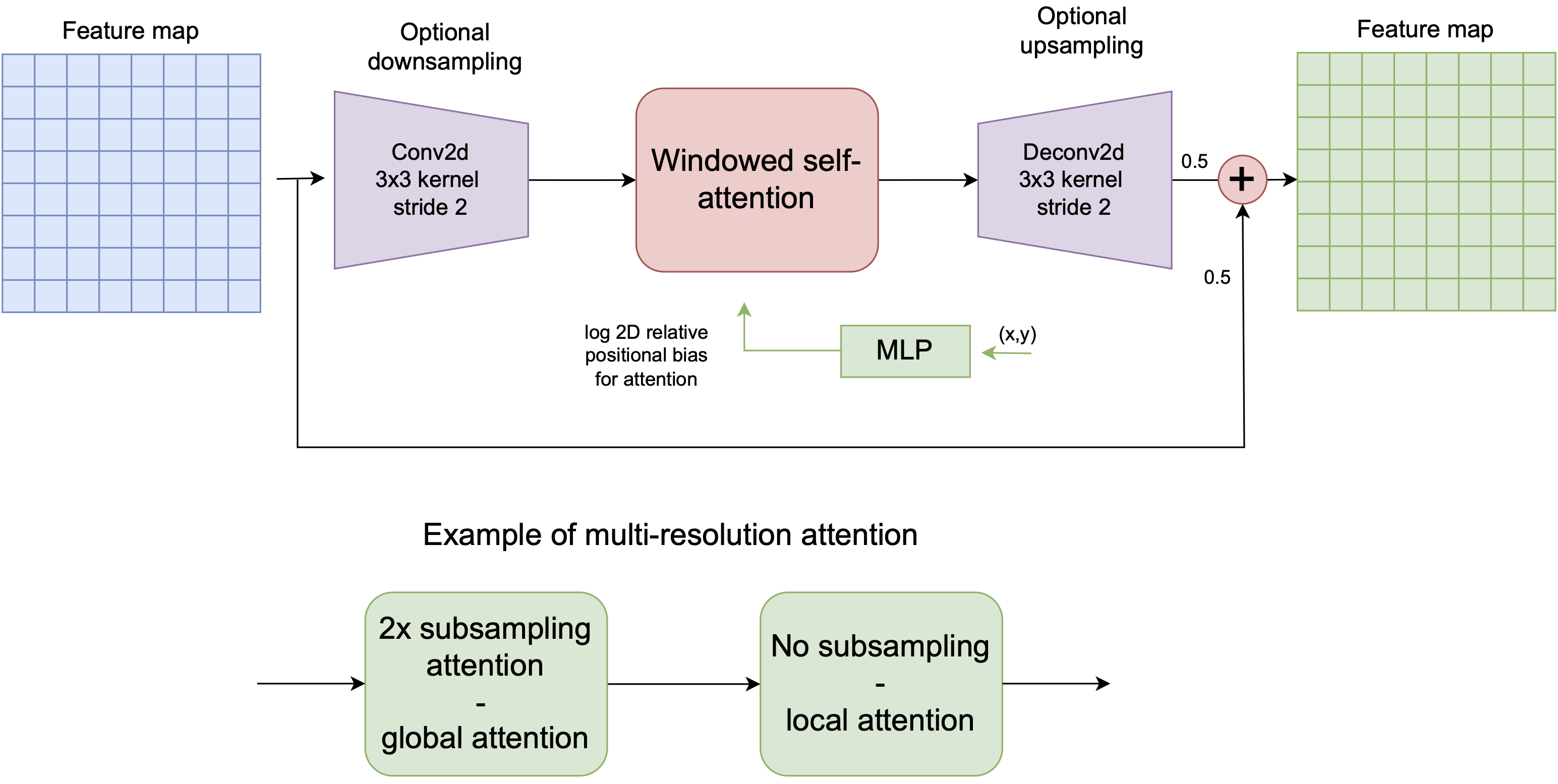}
    \caption{Multi-resolution attention for E-RADIO}
    \label{fig:eradio_mra}
\end{figure}

Standard transformers struggle to scale with high input image resolution because of quadratic complexity of the attention. SWIN~\cite{liu2022swin} proposed to use windowed attention to reduce the complexity of attention. We reuse windowed attention in the E-RADIO. To address for missing communication between windows, SWIN introduced window shifting, unfortunately, it has non-negligible compute cost. Instead, we propose multi-resolution attention inspired by EdgeViT's Local-Global-Local attention~\cite{edgevit}. The idea is illustrated in Figure \ref{fig:eradio_mra}. Every layer in the transformer will have a local windowed attention with optional subsampling via convolutional operator. For example, if susbampling is dissabled, then it is just a standard windowed attention. If the subsampling ratio is 2, then the feature map is downsampled by a factor of 2, windowed attention is performed, and then the feature map is upsampled to the original resolution with deconvolution. For FasterVIT2 models, we interleave subsampled attention with ratio 2 and the normal attention with no subsampling.

\subsection{Configurations}

All models in the family follow the same configuration except the embedding dimension (hide dimension). We simply scale it up with bigger models. Other parameters:
\begin{itemize}
    \item Input resolution is 224
    \item In-stem contains 2 3x3 convolutions with stride 2
    \item Total stages: 2 convolutional and 2 transformer
    \item First stage takes input feature size of 56x56, has 3 layers with C2f structure from YOLO8~\cite{yolov8_ultralytics}. 
    \item Second stage takes input feature size of 28x28, has 3 layers of C2f. 
    \item Third stage takes features of size 14x14, has 5x multi-resolution attention, window size 7.
    \item Forth stage takes features of size 7x7, has 5x windowed attention of window size 7.
    \item Embedding dimension for different model variants: XT - 64, T - 80, S - 96, B - 128, L - 192. The smallest XT and T models have [1, 3, 4, 5] layers for each of 4 stages. 
    \item Output features have resolution of 14x14 and are obtained by upsampling the features of stage 4 by 2x with deconvolution and adding to stage 3 features of size 14x14. 
\end{itemize}

\section{PCA Visualizations}\label{apdx:pca_viz}

We visualize various models using PCA to reduce the model's spatial feature dimensionality down to 3 dimensions, and directly map those to RGB. Most models are only able to handle square inputs at fixed resolutions, however DINOv2 and RADIO can handle arbitrary resolutions and aspect ratios, so we visualize them in both settings.

\subsection{Square Models}
\renewcommand{\arraystretch}{0}  
\begin{tabular}{cm{1.5cm}m{19cm}}

    \textbf{Model} & \textbf{Resolution} & \textbf{Images} \\
    & & \resizebox{0.5\linewidth}{!}{\includegraphics[]{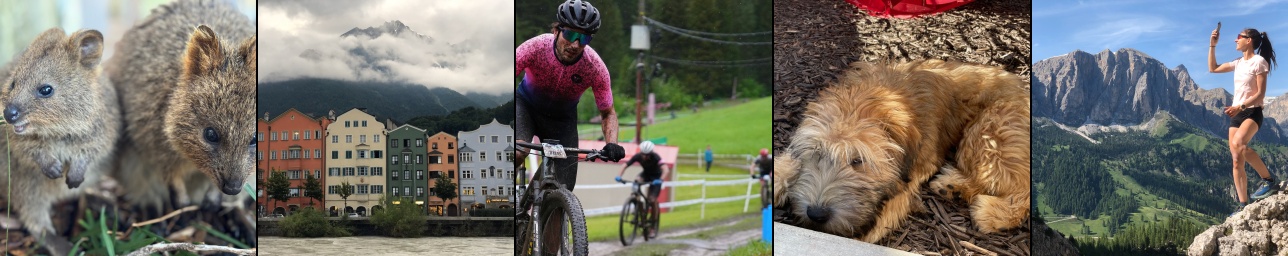}} \\
    OpenCLIP-H/14 & 224 & \resizebox{0.5\linewidth}{!}{\includegraphics[]{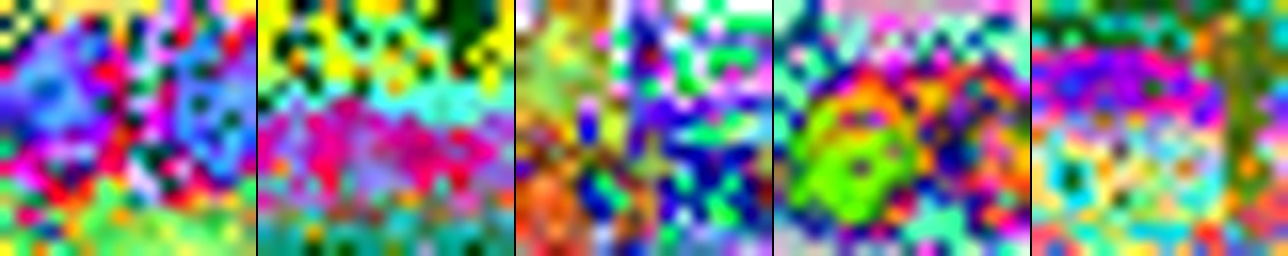}} \\
    MetaCLIP-H/14 & 224 & \resizebox{0.5\linewidth}{!}{\includegraphics[]{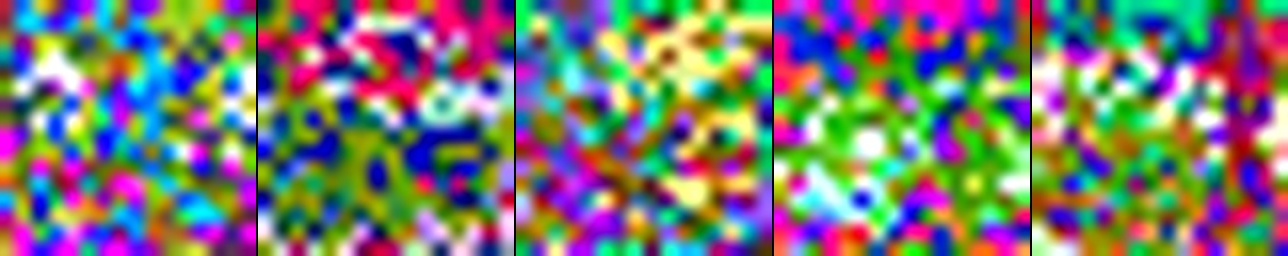}} \\
    SigLIP-M/14   & 384 & \resizebox{0.5\linewidth}{!}{\includegraphics[]{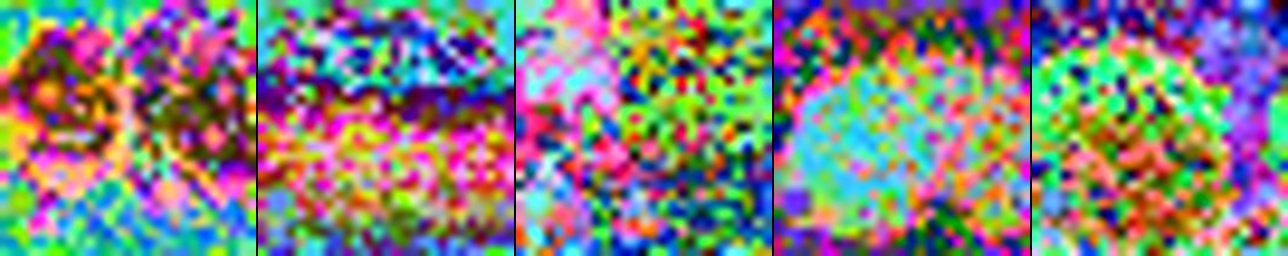}} \\
    InternViT-6B  & 224 & \resizebox{0.5\linewidth}{!}{\includegraphics[]{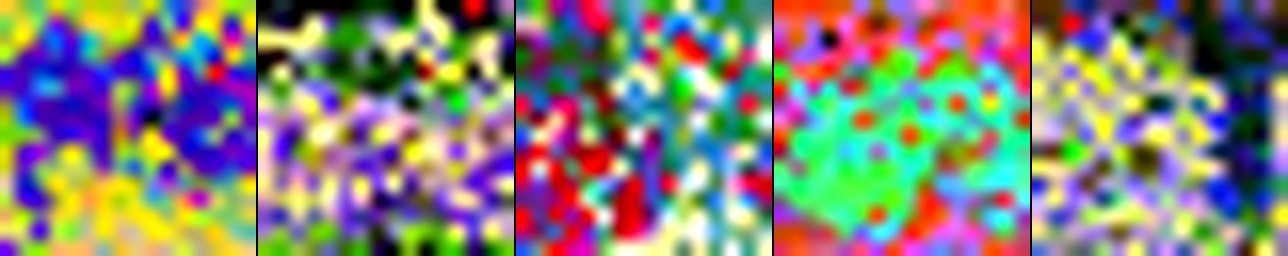}} \\
                  & 448 & \resizebox{0.5\linewidth}{!}{\includegraphics[]{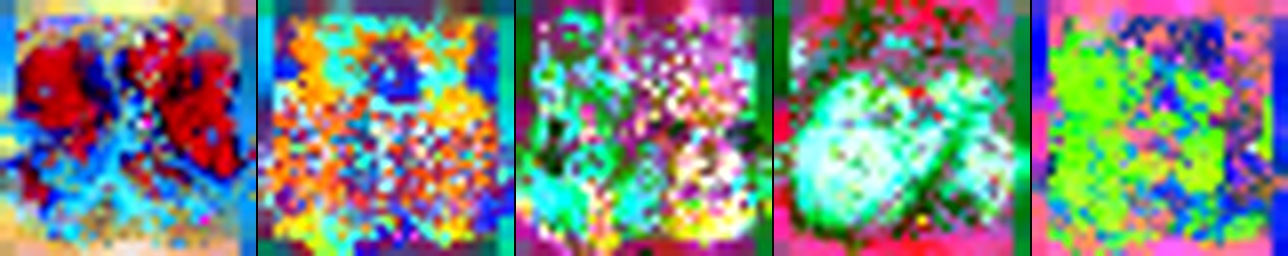}} \\
    DFN CLIP      & 378 & \resizebox{0.5\linewidth}{!}{\includegraphics[]{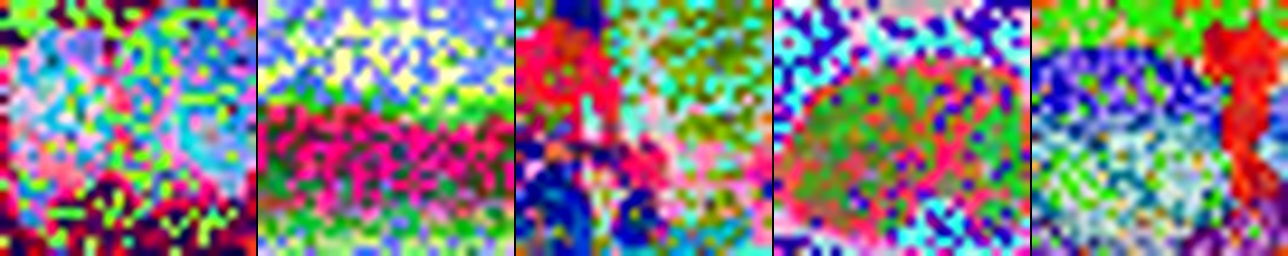}} \\
    OpenAI CLIP   & 336 & \resizebox{0.5\linewidth}{!}{\includegraphics[]{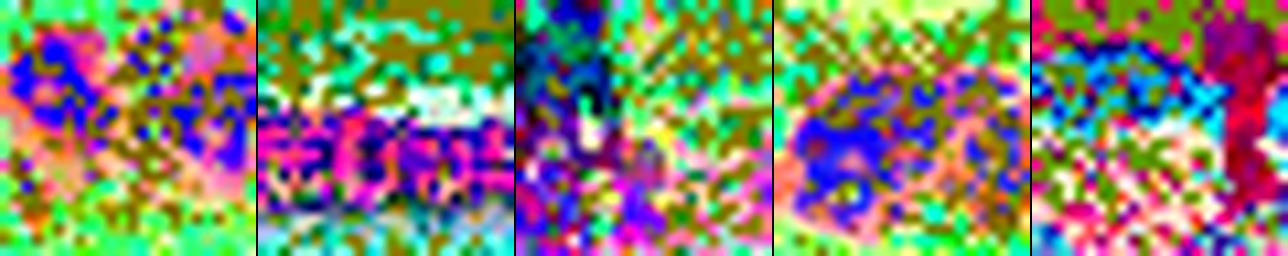}} \\
    DINOv2-g      & 518 & \resizebox{0.5\linewidth}{!}{\includegraphics[]{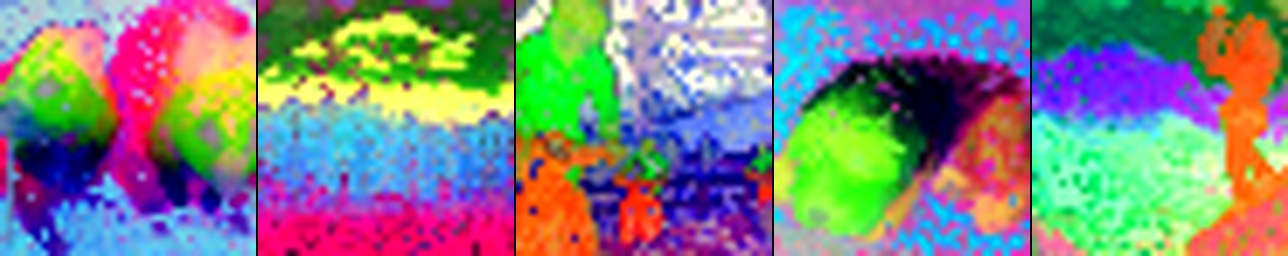}} \\
    SAM-H         & 1024 & \resizebox{0.5\linewidth}{!}{\includegraphics[]{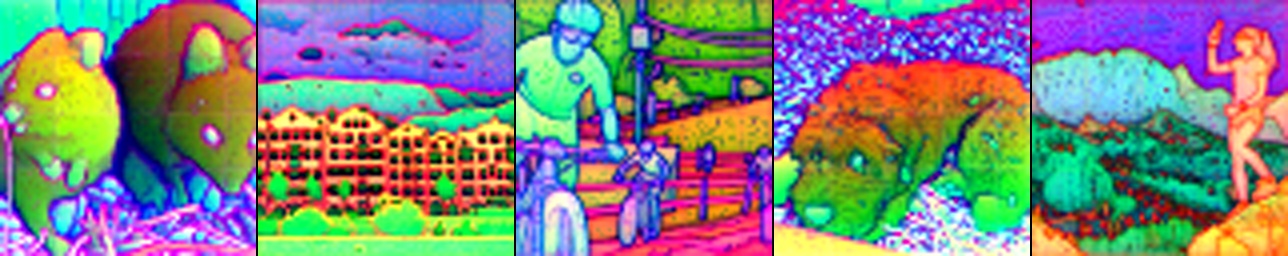}} \\
    RADIO         & 512 & \resizebox{0.5\linewidth}{!}{\includegraphics[]{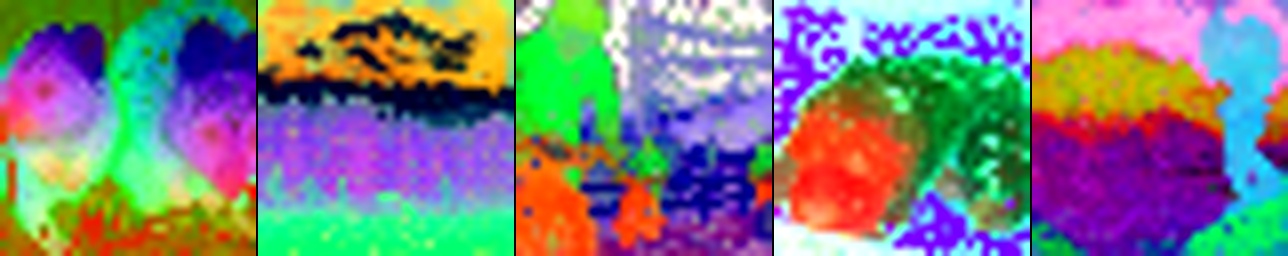}} \\
                  & 1024 & \resizebox{0.5\linewidth}{!}{\includegraphics[]{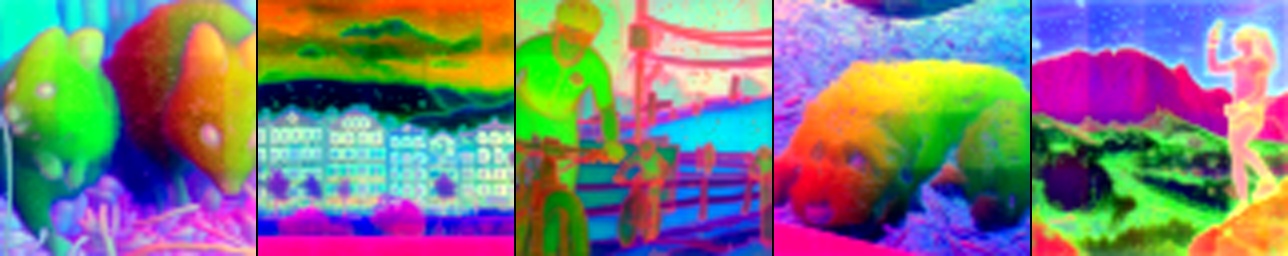}} \\
\end{tabular}
\renewcommand{\arraystretch}{1}  

\subsection{Flexible Models}

\renewcommand{\arraystretch}{0}  
\begin{tabular}{cm{1.5cm}m{24cm}}
    \textbf{Model} & \textbf{Resolution} & \textbf{Images} \\
    & & \resizebox{0.5\linewidth}{!}{\includegraphics[]{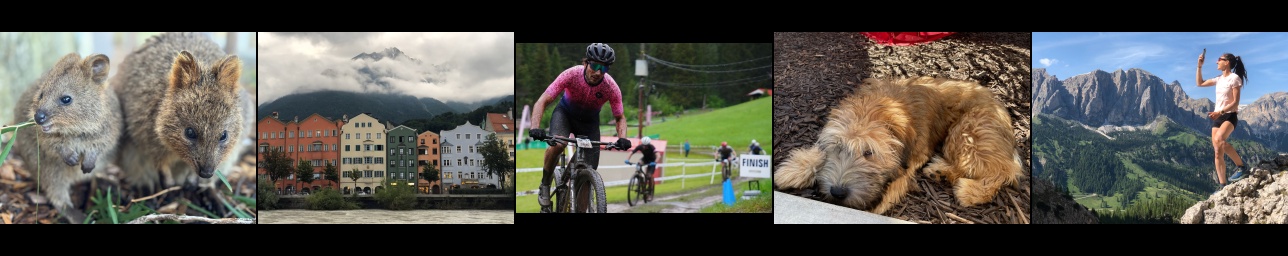}} \\
    DINOv2-g & 518  & \resizebox{0.5\linewidth}{!}{\includegraphics[]{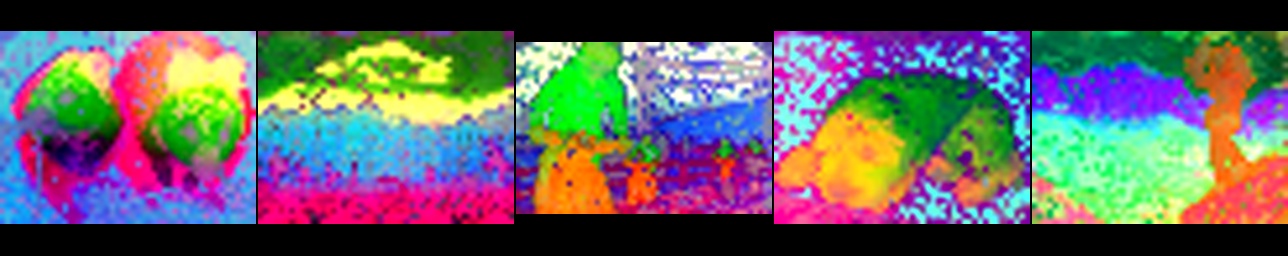}} \\
             & 1022 & \resizebox{0.5\linewidth}{!}{\includegraphics[]{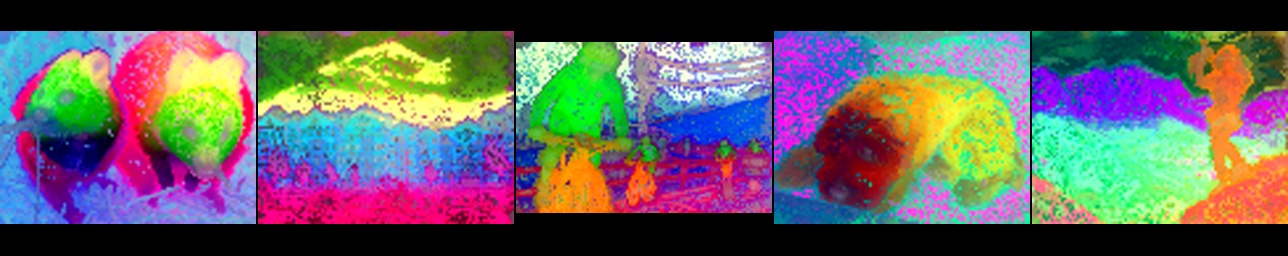}} \\
             & 2044 & \resizebox{0.5\linewidth}{!}{\includegraphics[]{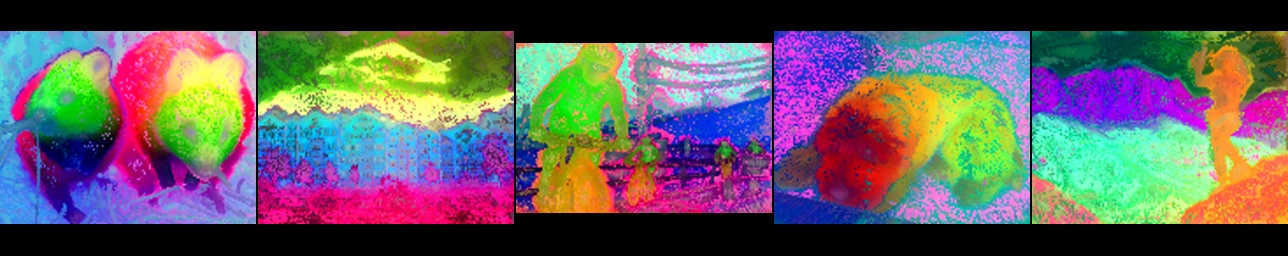}} \\
    RADIO    & 512  & \resizebox{0.5\linewidth}{!}{\includegraphics[]{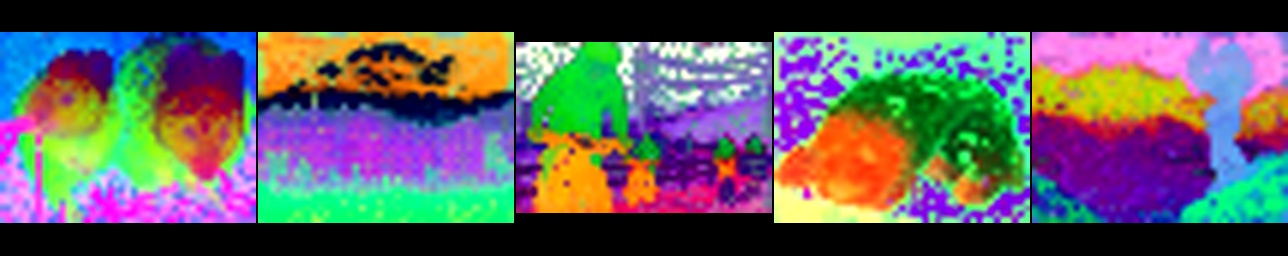}} \\
             & 1024 & \resizebox{0.5\linewidth}{!}{\includegraphics[]{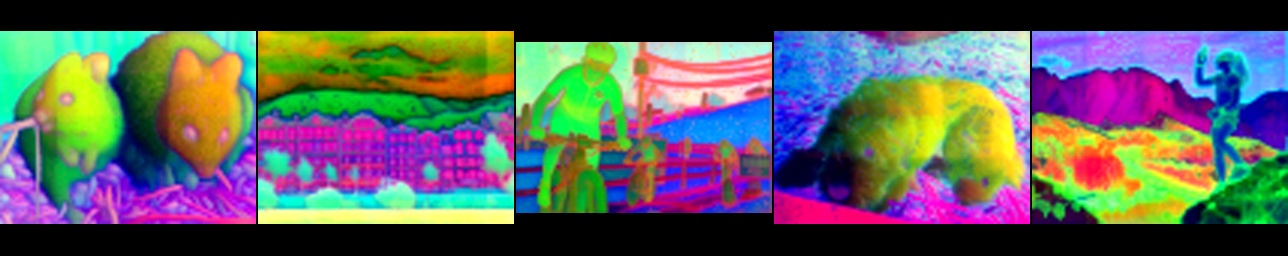}} \\
             & 2048 & \resizebox{0.5\linewidth}{!}{\includegraphics[]{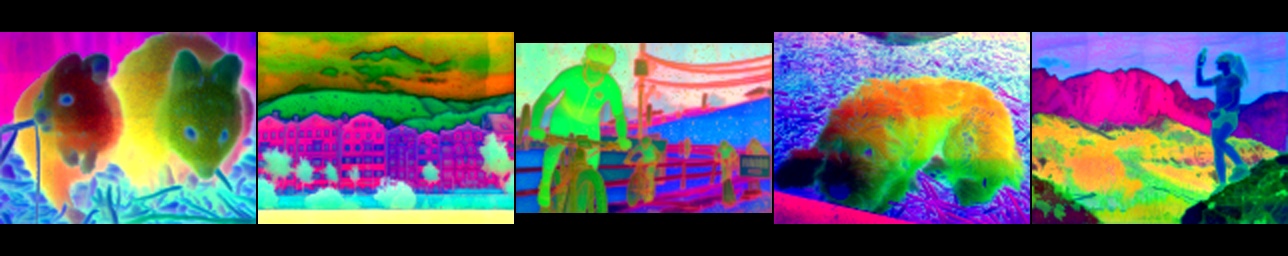}} \\
\end{tabular}
\renewcommand{\arraystretch}{1}  

\section{ViTDet Augmentation}\label{apdx:vitdet_aug}

The following python code shows how the alternating window/global architecture of ViTDet \cite{li2022vitdet} can be applied to a transformer. We take advantage of the fact that transformers are permutation invariant \textit{after position encodings have been applied}, and thus it's easy to organize the patch order such that contiguous chunks of patches belong to the same window. Once reordered in this way, alternating between windowed and global attention is achieved simply by absorbing the windows into the batch dimension or returning to the original shape respectively. We also enforce that the final transformer layer always applies global attention.

\begin{lstlisting}[language=Python, showstringspaces=false]
from einops import rearrange
def reorder_patches(patches: torch.Tensor, 
                    patched_size: Tuple[int, int], 
                    window_size: int):
    p_idxs = torch.arange(patches.shape[1])
    p_idxs = rearrange(p_idxs, '(wy y wx x) -> (wy wx y x)',
                       wy=patched_size[0] // window_size, y=window_size,
                       wx=patched_size[1] // window_size, x=window_size)
    p_idxs = p_idxs.reshape(1, -1, 1).expand_as(patches)
    
    return torch.gather(patches, p_idxs), p_idxs

def vitdet_aug(blocks: nn.Sequential,
               patches: torch.Tensor,
               patched_size: Tuple[int, int],
               window_sizes: List[int],
               num_windowed: int):
    B, T, C = patches.shape
    window_size = sample(window_sizes)
    sq_window_size = window_size ** 2
    patches, p_idxs = reorder_patches(patches, patched_size, window_size)
    period = num_windowed + 1
    for i, block in enumerate(blocks[:-1]):
        if i % period == 0:
            patches = patches.reshape(B * sq_window_size, -1, C)
        elif i % period == num_windowed:
            patches = patches.reshape(B, T, C)
        patches = block(patches)

    # Always use global attention with the last block
    patches = patches.reshape(B, T, C)
    patches = blocks[-1](patches)

    # Finally, put the patches back in input order
    ret = torch.empty_like(patches)
    ret = ret.scatter(dim=1, index=p_idxs, src=patches)
    return ret
\end{lstlisting}

\section{Comparison with SAM-CLIP \cite{wang2023samclip}}

Concurrently with our work, SAM-CLIP was introduced as a method of fusing SAM and CLIP into a single model. Due to the concurrency of effort, we don't compare our model with the full suite of metrics demonstrated in their method, however, we do have some overlap in key metrics such as Zero-Shot ImageNet-1k, and ADE20k semantic segmentation via linear probing. We present the comparison in table \ref{tab:comp_sam_clip}, however we note that there are enough differences between these two models that we can't conclude one way or another what is the superior approach. Instead we'll argue that DINOv2 does a better job of ADE20k linear probing than SAM, and thus our significantly higher quality on this metric is likely due to the inclusion of DINOv2, which is a key introduction with our approach.

\begin{table}[]
    \centering
    \begin{tabular}{r|c|c|c}
        \textbf{Family} & \textbf{Model} & \textbf{Zero-Shot} & \textbf{ADE20k} \\
        \hline
        SAM             & ViTDet-H/16    &                    & 28.2 \\
        DFN CLIP        & ViT-H/14       & \textbf{83.9}      & 31.7 \\
        \hline
        SAM-CLIP        & ViTDet-B/16    & 71.7               & 38.4 \\
        RADIO           & ViT-H/14       & 82.7               & \textbf{51.3}
    \end{tabular}
    \caption{We compare our common key metrics with those demonstrated in SAM-CLIP \cite{wang2023samclip}. We note that there are numerous differences between the two approaches, including model capacity and architecture. SAM-CLIP uses the ViT-B variant of SAM as a starting point, which implies it's a ViTDet-B/16 architecture. As a result of this choice, their metrics are computed at a resolution of 1024. RADIO trains a vanilla ViT-H/14 from scratch, and as a result of the flexibility gained via the CPE method, we evaluate Zero-Shot ImageNet1k at a resolution of 432, and we run ADE20k linear probing at a resolution of 512 using the exact same weights. We note that Zero-Shot quality is largely determined by the quality of the CLIP teacher and the capacity of the student. We attribute our superior quality on ADE20k semantic segmentation largely to our inclusion of DINOv2 as a teacher.}
    \label{tab:comp_sam_clip}
\end{table}

\section{Automatic Loss Balancing}\label{apdx:autobalance}
\subsection{Uncertainty}\label{apdx:uncertainty}

Following \cite{cipolla2018autobalance}, we have:

\begin{equation}
    L(x) = \sum_k \frac{1}{2\sigma_k^2} L_k(x) + \log \sigma_k
\label{eq:uncertainty_definition}
\end{equation}

where the $\sigma_k$ values are predicted by the student. In practice, the student predicts $b := \log \sigma_k^2$ for numerical stability, to avoid division by zero, and to regress unconstrained scalar values.

We make some minor modifications to \eqref{eq:uncertainty_definition} to make training a bit more stable in our setting. We replace the manual $\lambda$ scalars with the learned uncertainty weights, and add the loss term for large uncertainties. Altogether, this yields:

\begin{equation}
\begin{aligned}
    \lambda_k &= \frac{e^{-b_k}}{2} \\
    L(x)      &= \sum_k \lambda_k L_k(x) + \frac{b_k}{2}
\end{aligned}
\label{eq:uncertainty_definition_2}
\end{equation}

Let $b_i^{(s|v)}(x'|\Theta_i^{(s)})$ be a learned function predicting balance parameters for teacher $i$ and summary weight $(s)$ or feature vector weight $(v)$, we transform equation \eqref{eq:uncertainty_definition_2} slighty to:

\begin{equation}
\begin{aligned}
    \psi(x) &= \log (1 + e^x) \\
    \lambda_i^{(m)} &= e^{-b_i^{(m)}(x')} \\
    L(x) &= \sum_i \sum_{m \in \{s,v\}} \lambda_i^{(m)} L_i^{(m)}(x) + \psi\left(b_i^{(m)}(x')\right)
\end{aligned}
\label{eq:uncertainty_ours}
\end{equation}

The function $\psi(x)$ is the familiar ``softplus'' nonlinear activation function. We drop the division by 2 on the left because, assuming outputs are initially $b \sim \mathcal{N}(0, \sigma^2)$, then the loss weights will initially have an expected value of 1, matching the naive weighting. On the right, we replace $\frac{b_k}{2}$ with $\psi(x)$ for a few reasons:

\begin{itemize}
    \item When $x \gtrapprox 4$, then $\psi(x) \approx x$, yielding the same expression as before.
    \item When $x \approx 0$, then $\psi'(x) \approx \frac{1}{2}$, yielding the same expression as before.
    \item When $x < 0$, which translates to a loss weight $> 1$, $\psi'(x) \to 0$, improving stability as the weight gets larger.
    \item It has range $(0, \infty)$ which aesthetically enforces the loss to be greater than zero.
\end{itemize}

\subsection{AdaLoss}\label{apdx:adaloss}
In addition to uncertainty auto-balancing, we also explored AdaLoss \cite{hu2019adaloss}. In this formulation, we have:

\begin{equation}
\begin{aligned}
    \lambda_i^{(m)} &= \frac{1}{\mathbb{E}(L_i^{(m)})} \\
    L(x) &= \sum_i \sum_{m \in \{s,v\}} \lambda_i^{(m)} L_i^{(m)}(x)
\end{aligned}
\label{eq:adaloss}
\end{equation}

\section{Visual Question Answering Samples}

\begin{figure}
    \centering
    \includegraphics[width=0.9\linewidth]{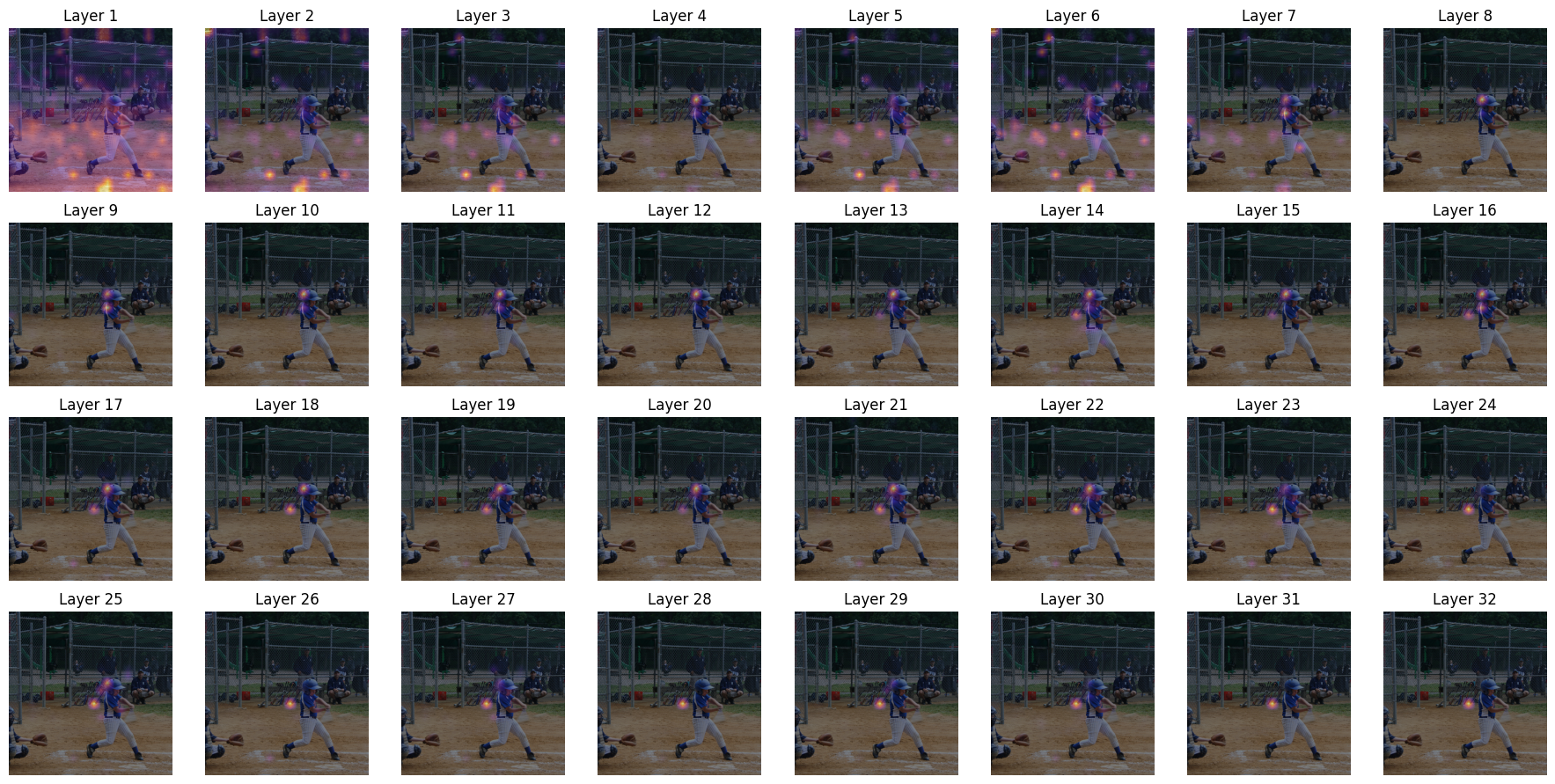}
    \caption{Visualization of the LLaVA attention maps over the visual features produced by a RADIO encoder. We use one sample image from the GQA\cite{DBLP:journals/corr/abs-1902-09506} validation set and one associated question: "What color is the helmet in the middle of the image?". For each layer in the language model, we retrieve attention scores for all positions of the visual tokens, average them over all attention heads, and overlay corresponding heat maps with the input image. We can see that as we progress through the layers, the model's attention focuses on the relevant part of the image. The model's answer is "Blue".}
    \label{fig:gqa_attention}
\end{figure}

\begin{figure}
    \centering
    \includegraphics[width=0.9\linewidth]{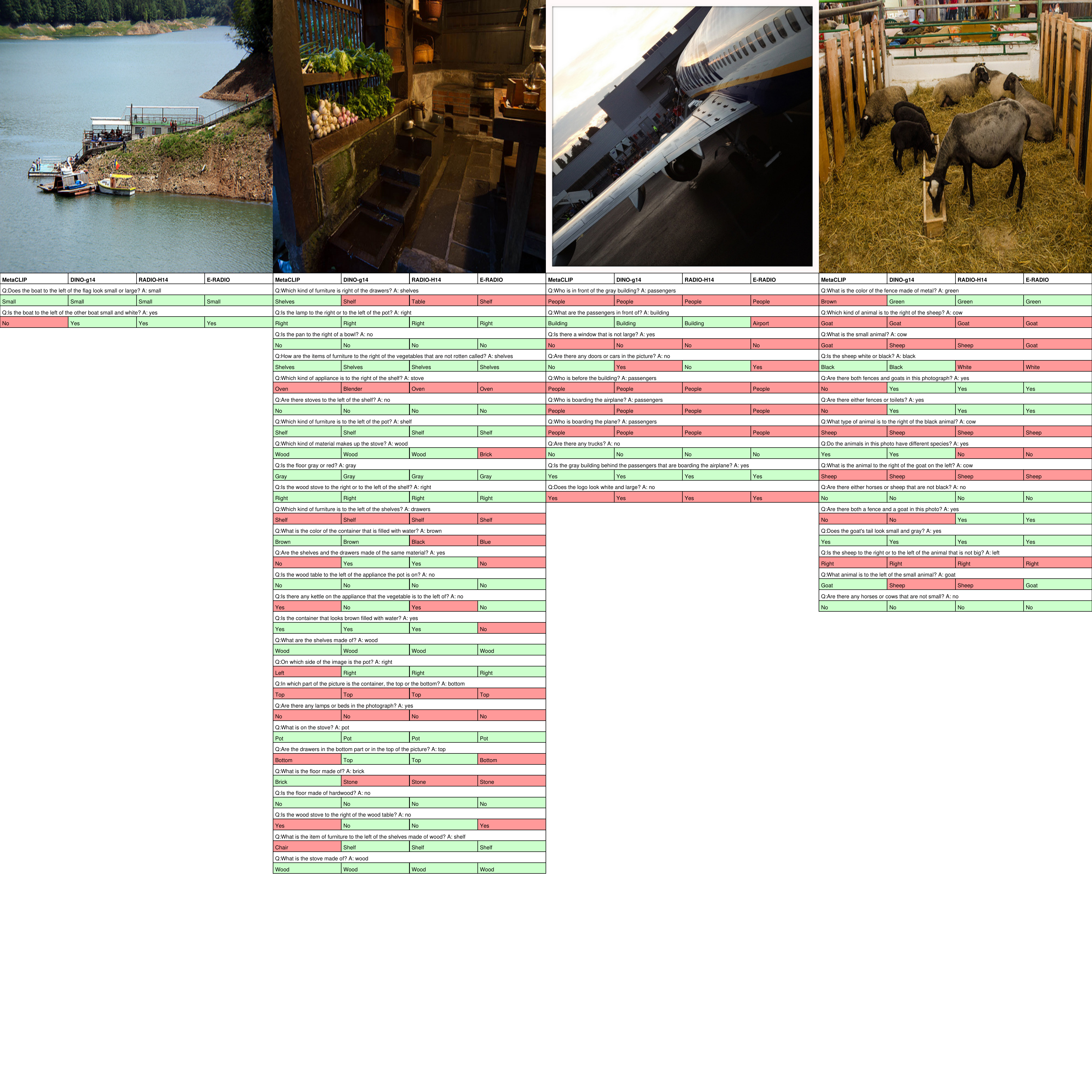}
    \caption{Sample questions from the GQA\cite{DBLP:journals/corr/abs-1902-09506} and their answers from our LLaVA models, using various image encoders. Answers are painted green when they match the ground truth, pink otherwise.}
    \label{fig:gqa_samples_1}
\end{figure}

\begin{figure}
    \centering
    \includegraphics[width=0.9\linewidth]{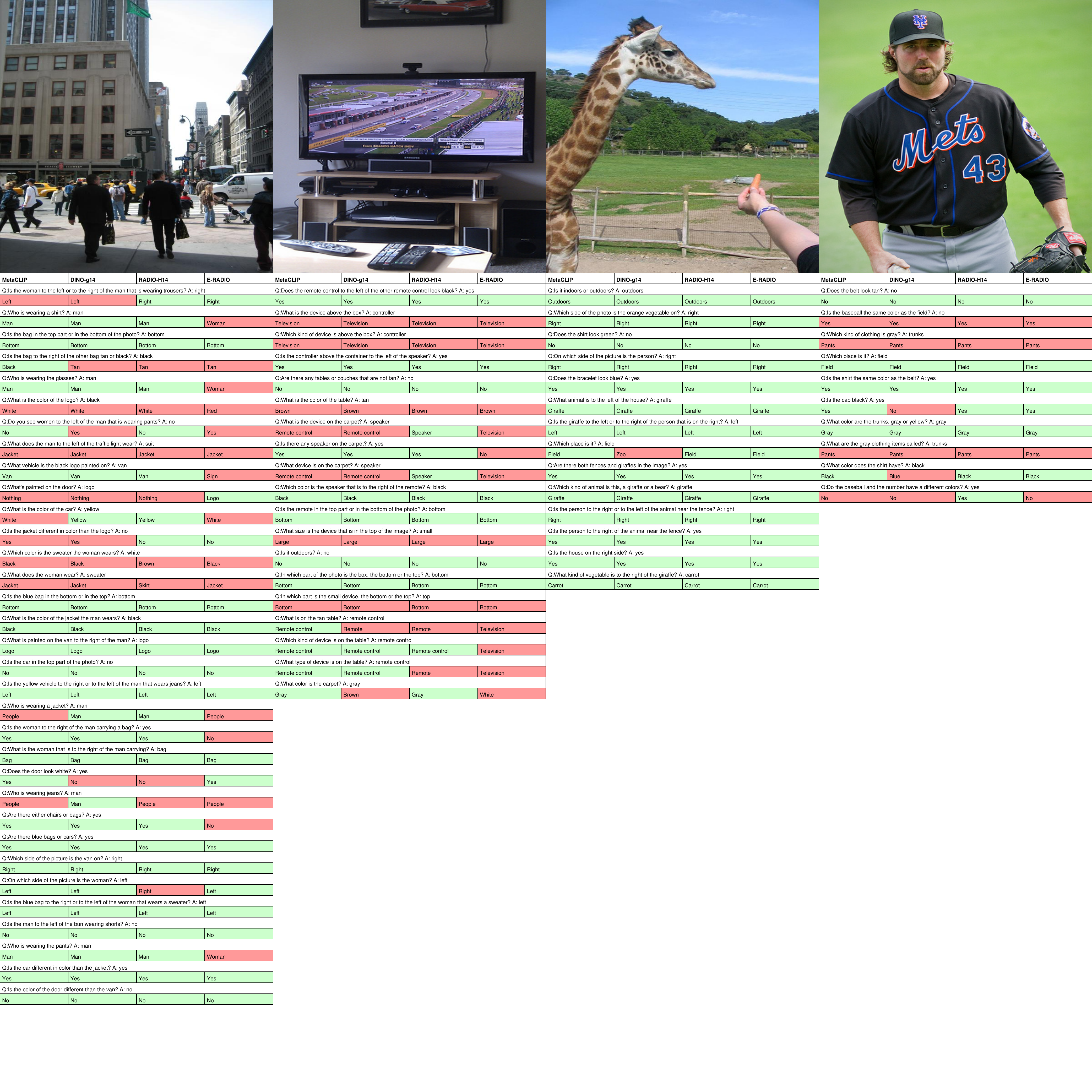}
    \caption{Sample questions from the GQA\cite{DBLP:journals/corr/abs-1902-09506} and their answers from our LLaVA models, using various image encoders. Answers are painted green when they match the ground truth, pink otherwise.}
    \label{fig:gqa_samples_2}
\end{figure}


\begin{figure}
    \centering
    \includegraphics[width=0.9\linewidth]{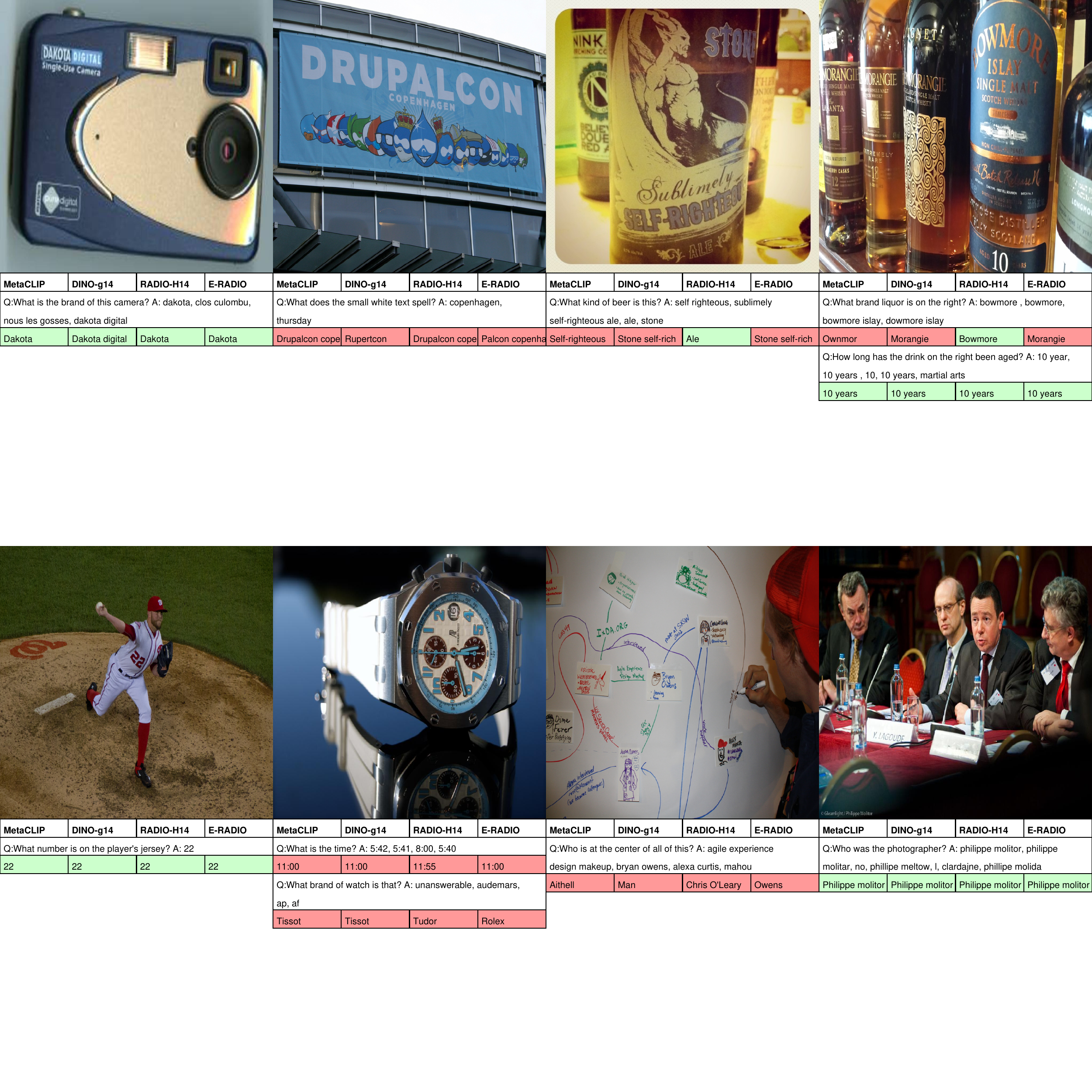}
    \caption{Sample questions from the TextVQA \cite{Singh_2019_CVPR} dataset and their answers from our LLaVA models, using various image encoders. Answers are painted green when they match the ground truth, pink otherwise.}
    \label{fig:textvqa_samples}
\end{figure}

\begin{figure}
    \centering
    \includegraphics[width=0.9\linewidth]{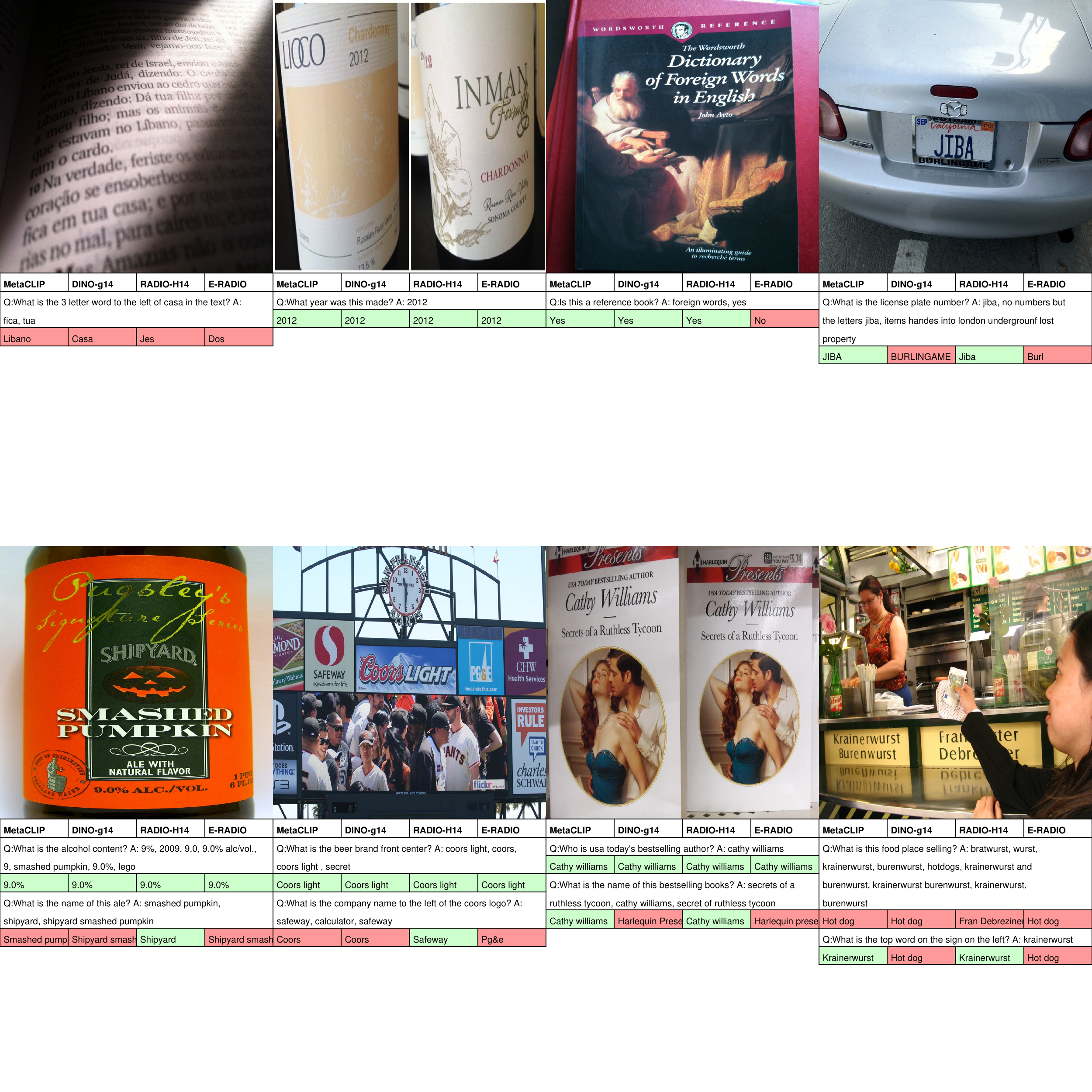}
    \caption{Sample questions from the TextVQA \cite{Singh_2019_CVPR} dataset and their answers from our LLaVA models, using various image encoders. Answers are painted green when they match the ground truth, pink otherwise.}
    \label{fig:textvqa_samples_2}
\end{figure}

\begin{figure}
    \centering
    \includegraphics[width=0.9\linewidth]{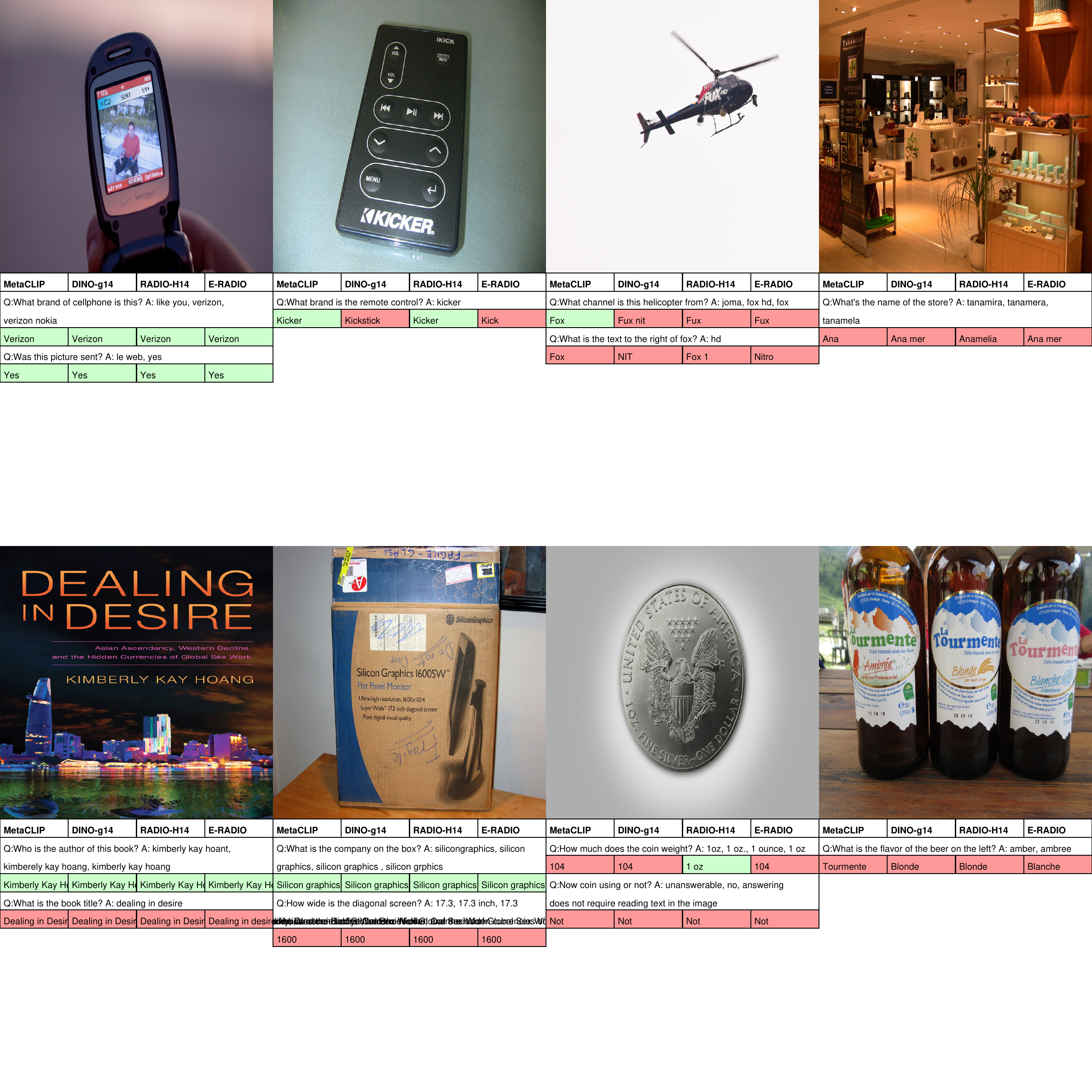}
    \caption{Sample questions from the TextVQA \cite{Singh_2019_CVPR} dataset and their answers from our LLaVA models, using various image encoders. Answers are painted green when they match the ground truth, pink otherwise.}
    \label{fig:textvqa_samples_3}
\end{figure}

Figures \ref{fig:gqa_samples_1} to \ref{fig:textvqa_samples_3} show sample questions from our Visual Question Answering datasets, together with sample answers when using our vision encoders in a LLaVA setup.